\documentclass[acmtog]{acmart}

\usepackage{booktabs} 
\usepackage{cleveref}
\usepackage{multirow}

\usepackage[ruled]{algorithm2e} 

\SetAlFnt{\small}
\SetAlCapFnt{\small}
\SetAlCapNameFnt{\small}
\SetAlCapHSkip{0pt}

\begin{document}
\title{TOPOS: High-Fidelity and Efficient Industry-Grade 3D Head Generation}

\author{Bojun Xiong}
\authornote{Equal contribution}
\author{Zoubin Bi}
\authornotemark[1]
\author{Xinghui Peng}
\author{Yunmu Wang}
\authornote{Project lead}
\author{Junchen Deng}
\author{Jun Liang}
\author{Jing Li}
\author{Bowen Cai}
\author{Huan Fu}
\authornote{Corresponding author}
\affiliation{
  \institution{HUJING Digital Media \& Entertainment Group}
  \city{Beijing}
  \country{China}
}


\newcommand{\figref}[1]{Fig.~\ref{#1}}
\newcommand{\tabref}[1]{Tab.~\ref{#1}}
\newcommand{\secref}[1]{Sec.~\ref{#1}}
\newcommand{\note}[1]{{\color{blue}#1}}
\newcommand{\warning}[1]{{\color{red}#1}}

\begin{abstract}

High-fidelity 3D head generation plays a crucial role in the film, animation and video game industries. In industrial pipelines, studios typically enforce a fixed reference topology across all head assets, as such a clean and uniform topology is a prerequisite for production-level rigging, skinning and animation. In this paper, we present TOPOS, a framework tailored for single image conditioned 3D head generation that jointly recovers geometry and appearance under such an industry-standard topology. In contrast to general 3D generative models which produce triangle meshes with inconsistent topology and numerous vertices, hindering semantic correspondence and asset-level reuse, TOPOS generates head meshes with a fixed, studio-style topology, enabling consistent vertex-level correspondence across all generated heads. To model heads under this unified topology, we proposed a novel variational autoencoder structure, termed TOPOS-VAE. Inspired by multi-model large language models (MLLMs), our TOPOS-VAE leverages the Perceiver Resampler to convert input pointclouds sampled from head meshes of diverse topologies into the target reference topology. Building upon TOPOS-VAE's structured latent space, we train a rectified flow transformer, TOPOS-DiT, to efficiently generate high-fidelity head meshes from a single image. We further present TOPOS-Texture, an end-to-end module that produces relightable UV texture maps from the same portrait image via fine-tuning a multimodal image generative model. The generated textures are spatially aligned with the underlying mesh geometry and faithfully preserve high-frequency appearance details. Extensive experiments demonstrate that TOPOS achieves state-of-the-art performance on 3D head generation, surpassing both classical face reconstruction methods and general 3D object generative models, highlighting its effectiveness for  digital human creation. We will release our code and trained models to facilitate future research.
\end{abstract}

\begin{teaserfigure}
  \centering
  \includegraphics[width=\textwidth]{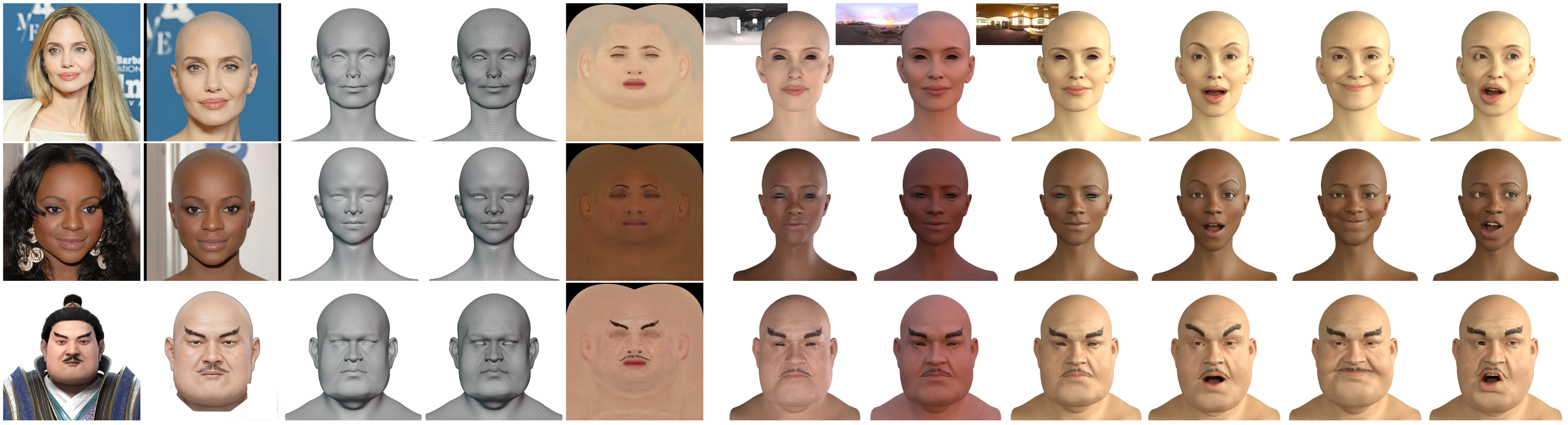}
  \vspace{-2mm}
  \caption{Our proposed TOPOS framework is capable of generating high-fidelity 3D head mesh and texture map given a single image. From left to right are the input images, edited images, generated geometry, mesh topology, generated texture maps, rendering images under different lighting conditions and animation results across different expressions, respectively. The upper-left insets illustrate the environment maps. Please zoom in for better inspection.}
  \label{fig:teaser}
\end{teaserfigure}

%
%

\ccsdesc[500]{Computing methodologies~Mesh models}
\ccsdesc[500]{Computing methodologies~Texturing}

%
%

\keywords{3D Head Generation, Geometry Modeling, Texture Modeling}

\maketitle

\vspace{-4mm}
\section{Introduction}
High-fidelity 3D head generation is of vital importance in the film, computer animation and video game industries. However, it is still quite time-consuming for a skilled artist using professional tools to create a realistic and industry-grade head asset, which is a labor-intensive process taking several hours or days per head. Moreover, industrial production pipelines typically enforce a fixed studio-defined reference topology across all head assets within a project, since a clean and uniform topology is a prerequisite for production-level rigging, skinning and downstream character animation. Therefore, developing an automatic algorithm for high-fidelity and efficient 3D head generation with a uniform topology and relightable texture map would be highly meaningful and important in the field of Computer Graphics and digital human creation.

Traditional 3D face reconstruction methods utilize parametric face model, such as 3D Morphable Models (3DMMs)~\cite{3dmm1999} to represent the 3D face mesh. 3DMMs provide a compact and controllable representation of facial geometry and appearance, enabling stable reconstruction from conditional input. Despite advances in dataset scale and model capacity~\cite{paysan20093d, FLAME2017}, their inherently limited parameterization makes it difficult to capture high-frequency details, which constrains the representation of fine-scale geometry and texture.

On the other hand, driven by the rapid development of deep generative models, particularly Diffusion Models~\cite{ho2020denoising} and Flow Matching~\cite{lipman2022flow, lipman2024flow}, extensive efforts have been devoted to 3D shape modeling using various representations, including point clouds~\cite{luo2021diffusion, vahdat2022lion}, Signed Distance Functions~\cite{zheng2023locally, xiong2025octfusion} and Flexicubes~\cite{xiang2025structured, he2025sparseflex}. While these representations enable fine-grained geometric modeling, they typically require an additional surface extraction step such as Marching Cubes algorithm~\cite{lorensen1998marching}, which often produces head meshes with excessive vertices and unstructured connectivity. Recent auto-regressive mesh generation methods~\cite{chen2024meshxl, chen2025meshanything, siddiqui2024meshgpt} avoid this indirect extraction but are constrained by the prohibitive cost of Transformer self-attention~\cite{vaswani2017attention}, limiting the number of generated vertices and faces. More importantly, none of these methods enforce a consistent topology across generated instances, restricting their applicability to downstream tasks, such as skeletal rigging and character animation.

For facial appearance generation, it is standard to use 2D texture maps to capture fine-scale details. However, inherent self-occlusions make the reconstruction of complete, hole-free textures challenging, even with dense multi-view inputs~\cite{lattas2021avatarme++, han2024high, han2025facial}. Prior methods address missing regions through template-based completion~\cite{bai2023ffhq} or UV-space inpainting~\cite{zeng2022joint, lei2023hierarchical, yang2025freeuv, qiu2025avatartex}, but these approaches often introduce seams and identity drift. In addition, inaccuracies in the unprojection process further degrade texture quality, especially in detail-sensitive regions. Recent end-to-end approaches such as Uni-1~\cite{luma_uni1} are closest in spirit to ours, but lack explicit semantic alignment, leading to inconsistent texture representations.

In this paper, we propose TOPOS, a well designed generative framework tailored for industry-grade 3D head generation. Our TOPOS framework consists of three separate modules, TOPOS-VAE, TOPOS-DiT and TOPOS-Texture for both geometry and texture generation. However, 3D head mesh datasets that conform to a unified industry-standard topology are inherently limited in scale and diversity. Therefore, beyond learning a continuous latent space for compact head mesh encoding, our TOPOS-VAE is also capable of converting pointclouds sampled from head meshes with diverse topologies into head meshes with the fixed and consistent reference topology to alleviate the scarcity of standardized head mesh data. Specifically, we adopt the Perceiver Resampler~\cite{alayrac2022flamingo}, which is widely used in multi-modal large language models (MLLMs), as the pointclouds encoder and Graph Neural Network (GNN) as the head mesh decoder. Leveraging its proven cross-modal alignment capability~\cite{alayrac2022flamingo}, Perceiver Resampler effectively translates unstructured point cloud features into structured mesh representations, which allows the GNN decoder to efficiently generate meshes with fixed topology by modeling mesh vertex connectivity as a graph structure. Building upon TOPOS-VAE's continuous and compact latent space, we train our head mesh generative model, termed TOPOS-DiT, a rectified flow transformer with rendered portrait images as condition. We further propose TOPOS-Texture, a dedicated end-to-end texture generation pipeline that produces relightable, geometry-consistent UV texture maps from the same single portrait image by leveraging rich identity and appearance priors from a pretrained multimodal image generative model (Qwen-Image-Edit~\cite{wu2025qwenimagetechnicalreport}). Benefiting from the compactness of TOPOS-VAE's latent space, TOPOS-DiT generates a head mesh in about one second. TOPOS-Texture is executed in parallel and is bottlenecked only by its backbone (around one minute in our implementation). As a result, our TOPOS framework is the first to efficiently generate high-fidelity 3D head assets that simultaneously preserve facial identity and conform to an industry-standard topology, significantly surpassing previous face reconstruction and 3D generation methods.
\section{Related Work}

\subsection{3D Face Reconstruction}

Parametric face models have long served as a foundational tool for 3D face reconstruction. The 3D Morphable Model (3DMM)~\cite{3dmm1999} parameterizes facial geometry and texture within a PCA space, and subsequent works extend this formulation through larger-scale datasets~\cite{paysan20093d,FLAME2017,booth2018large}, unconstrained imagery~\cite{kemelmacher2013internet, booth20173d, feng2021learning} and enhanced controllability over pose and expression~\cite{chai2022realy, FLAME2017, ploumpis2019combining, xu2020ghum, yang2020facescape}. Although these approaches naturally provide a fixed and consistent topology, their low-dimensional linear subspaces limit fine-grained geometric expressiveness. Recent rendering-based methods improve identity consistency via inverse rendering~\cite{zielonka2022towards}, neural image synthesis~\cite{retsinas2024smirk} or dense UV-space predictions~\cite{zeng2023flowface, giebenhain2025pixel3dmm}, yet still rely on a predefined parameter space.

Beyond parametric models, another line of work directly reconstructs 3D face geometry using vertex regression~\cite{richardson2017learning,sela2017unrestricted,zeng2019df2net} or neural volumetric representations such as NeRF~\cite{mildenhall2020nerf, wang2021neus}, Tri-plane~\cite{chan2022efficient} and 3DGS~\cite{kerbl3Dgaussians}. Recent methods~\cite{hu2024gaussianavatar,chu2024gagavatar, qiu2025lhm, wu2025fastavatar, li2021topologically, bolkart2023instant} achieve remarkable rendering realism, but typically output volumetric or point-based representations that are incompatible with industrial production pipelines. Therefore, existing 3D face reconstruction methods struggle to simultaneously achieve high-fidelity geometry and an industry-standard, uniform topology.

\subsection{3D Shape and Mesh Generation}

Deep generative models~\cite{ho2020denoising, song2020denoising, nichol2021improved, lipman2022flow} have rapidly advanced 3D shape generation across diverse representations, including implicit fields~\cite{mescheder2019occupancy, park2019deepsdf, deng2021deformed, zheng2023locally}, point clouds~\cite{nichol2022point, luo2021diffusion} and sparse voxel structures~\cite{xiong2025octfusion, he2025sparseflex, li2025sparc3d}. To enable scalable generation, various latent formulations have been explored, notably Perceiver-style unstructured latents~\cite{jaegle2021perceiver, zhang20233dshape2vecset, zhao2023michelangelo, zhao2025hunyuan3d} and structured latents~\cite{xiang2025structured, xiang2025trellis2}. While these methods can produce geometrically accurate 3D heads, the resulting meshes often contain excessive vertices and irregular connectivity, making them unsuitable for industrial production.

On the other hand, direct mesh generation methods aim to construct polygonal meshes with explicit topology in an end-to-end manner. Auto-regressive approaches such as MeshGPT~\cite{siddiqui2024meshgpt} and its variants~\cite{chen2024meshxl,weng2024pivotmesh} generate coherent meshes with regular face connectivity and subsequent works further improve mesh token compression~\cite{tang2024edgerunner, chen2025meshanything, song2025mesh}, tokenization strategies~\cite{lionar2025treemeshgpt, weng2025scaling, wang2025nautilus, liu2025freemesh} and model architectures~\cite{hao2024meshtron, wang2025iflame, fang2025meshllm, wang2024llama}. Although these methods can produce artist-style head meshes, they do not explicitly enforce a fixed topology across instances, which is incompatible with the uniform, reusable topology required by industrial production pipelines.

\vspace{-2mm}
\subsection{3D Face Texture Generation}

Early face reconstruction methods represent appearance using 3DMM texture coefficients or per-vertex colors, which cannot capture high-frequency facial details. Subsequent work moves to canonical UV-space recovery and optimizes texture, albedo or reflectance maps directly through inverse rendering~\cite{lattas2021avatarme++, han2024high, han2025facial}. While these optimization-based approaches can produce high-quality, relightable assets, they typically rely on multi-view observations, controlled illumination or short capture sequences, which limits their practicality in everyday settings.

To relax such requirements, recent methods address single-view texture reconstruction with learned priors. One line of work~\cite{bai2023ffhq, zeng2022joint, lei2023hierarchical, li2024uv, yang2025freeuv, qiu2025avatartex, dai2025high} employs generative models for novel-view synthesis or UV-space inpainting to complete self-occluded regions, while another line~\cite{wang2019digital,lattas2021avatarme++,dib2024mosar} directly regresses UV-related representations such as position or aligned texture maps to avoid explicit unprojection artifacts. These methods, however, still struggle in unobserved regions such as the inner mouth and often suffer from seams or local misalignment around detail-sensitive areas. More recently, large generative models have pushed the task toward end-to-end texture synthesis; the concurrent work Uni-1~\cite{luma_uni1} is closest to ours in this direction, but is trained in a straightforward manner on off-the-shelf datasets and lacks explicit semantic alignment with the underlying mesh layout. In contrast, our method explicitly enforces semantic alignment, producing UV textures that are more consistent and suitable for downstream avatar applications.
\section{Method}

\begin{figure*}[t!]
    \includegraphics[width=0.95\textwidth]{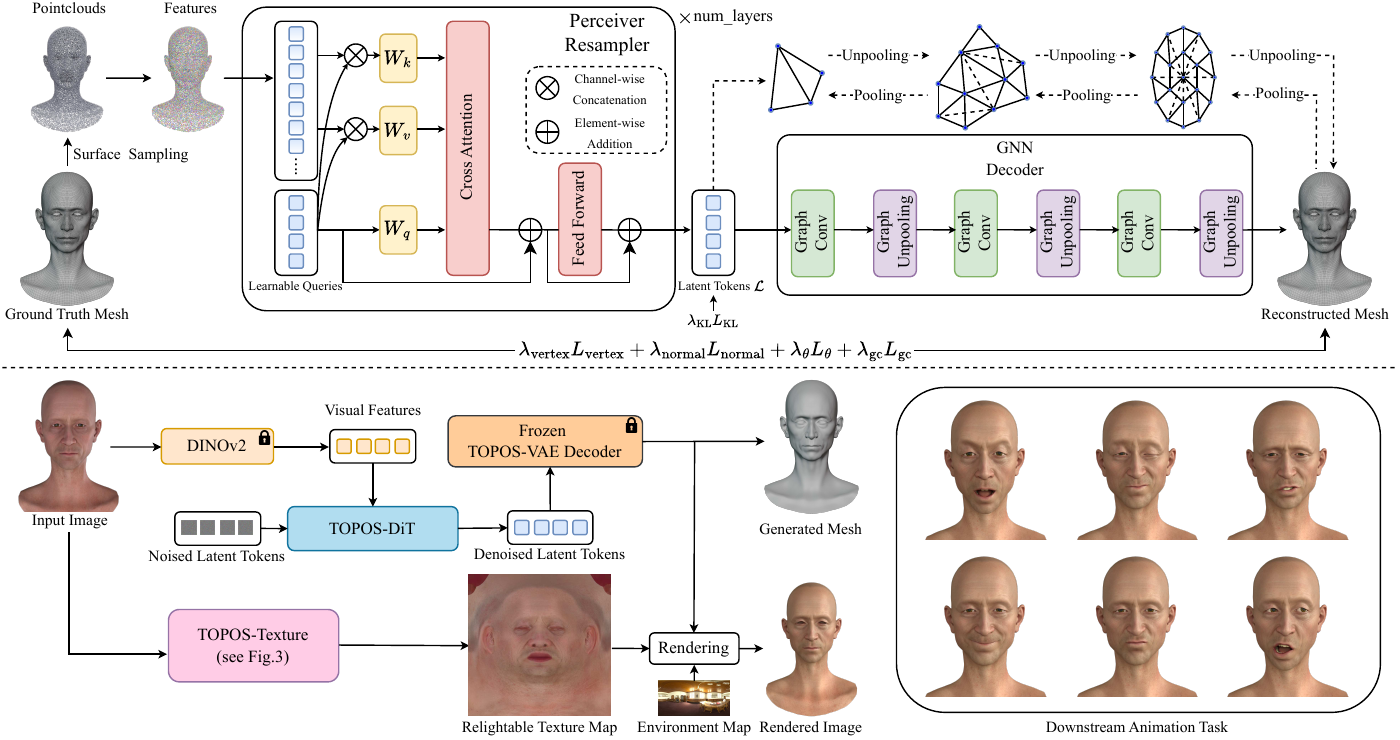}
    \vspace{-3mm}
    \caption{The top part of this figure shows the structure of our TOPOS-VAE, which utilizes Perceiver Resampler to encode the input pointclouds and GNN decoder to decode it into 3D head mesh with consistent topology. The bottom part shows the training and generation process of our TOPOS-DiT and TOPOS-Texture. Thanks to the unified topology, the generated 3D head mesh can be readily driven across different facial expressions for downstream applications. }
    \label{fig:method}
    \vspace{-2mm}
\end{figure*}

In this section, we provide a detailed explanation of our industry-grade 3D head generation framework, TOPOS. Specifically, we first train TOPOS-VAE to encode input pointclouds sampled from head meshes with diverse topologies into a continuous and compact latent space, which is decoded into 3D head meshes under a fixed, industry-standard topology to alleviate the scarcity of standardized head mesh datasets. Building upon this structured latent space, we subsequently train a rectified flow transformer, TOPOS-DiT, to model this latent distribution conditioned on a single rendered image. We further fine-tune a multimodal image generative model~\cite{wu2025qwenimagetechnicalreport} to generate a relightable UV texture map from the same input portrait. ~\figref{fig:method} shows an overview of our TOPOS framework, with details described below.

\subsection{TOPOS-VAE}

\subsubsection{Encoder}
A widely adopted design for 3D shape VAEs~\cite{zhao2025hunyuan3d, zhao2023michelangelo} is to employ the encoder from 3DShape2VecSet~\cite{zhang20233dshape2vecset}, which stacks cross-attention layers to encode input point clouds $\mathcal{P}$ using either learnable queries or farthest-point-sampled subsets as cross-attention queries. However, we argue that neither choice is suitable for our task. The latent space of TOPOS-VAE should form a set of \textbf{semantic anchors}, where each latent token corresponds to a specific semantic region on the fixed-topology head mesh. Learnable queries in 3DShape2VecSet struggle to establish such anchors due to the lack of iterative refinement with the input, leading to unstable training. On the other hand, farthest-point-sampled (FPS) queries, cannot guarantee consistent semantic correspondence across instances. As a result, 3DShape2VecSet-style encoders perform poorly on head mesh reconstruction under our fixed-topology setting in our pilot experiments, which is further demonstrated in the ablation studies.

Instead, inspired by the multi-modal large language models (MLLMs), we adopt the Perceiver Resampler~\cite{alayrac2022flamingo} as our encoder $\mathcal{E}$. The Perceiver Resampler maintains a set of learned queries $\mathcal{L} \in \mathbb{R}^{N \times d}$ shared across all instances. Unlike 3DShape2VecSet, where queries interact with the input only once via a single cross-attention layer, the Perceiver Resampler iteratively refines its queries by simultaneously attending to both the input pointcloud features and one another across multiple layers. This iterative refinement enables each query to specialize in a consistent semantic region across all training instances, naturally forming the desired semantic anchors. Since the queries are input-agnostic at initialization, their semantic roles emerge entirely through self-supervised learning, ensuring cross-instance consistency. The encoding process is formulated as $\mathcal{L} = \mathcal{E}(\mathcal{P}).$

\vspace{-1mm}
\subsubsection{Decoder}
We employ a Graph Neural Network (GNN) as our decoder $\mathcal{D}$, which takes latent tokens $\mathcal{L}$ from the Perceiver Resampler and outputs head meshes with consistent topology. Given the shared face connectivity $\mathcal{F}$ across all training meshes, the mesh structure naturally defines a graph $\mathcal{G}$.

\vspace{-2mm}
\paragraph{Graph Convolution} For vertex $v_i$ with feature $F_i$ and neighbors $\mathcal{N}(i)$, our graph convolution is defined as:
\begin{equation}
F_i' = W_0 F_i + \textstyle \sum_{j \in \mathcal{N}(i)} W_1 F_j,
\end{equation}
where $W_0$ and $W_1$ are trainable weights. This formulation is similar to GraphSAGE~\cite{hamilton2017inductive} and is simpler yet more effective than previous graph convolutions in 3D deep learning~\cite{wang2019dynamic} that define update functions as MLPs.

\vspace{-2mm}
\paragraph{Graph Pooling \& Unpooling} In our TOPOS-VAE, since GNNs are only employed in the decoder, the graph pooling operation is simply introduced to establish the mapping relation between vertices before and after pooling. Then we keep track of this mapping during graph unpooling in the decoder. 

We adopt grid-based graph pooling~\cite{simonovsky2017dynamic, pang2023learning} for efficiency. Grid-based pooling employs regular voxel grids to cluster vertices within the same voxel into a single vertex, constructing a hierarchy of graph structures at multiple levels:
\begin{equation}
\mathcal{G}^{(0)} \sim \{ \mathcal{V}_t, \mathcal{F} \}, \quad
\mathcal{G}^{(l+1)} = \mathrm{GridPool}^{(l)}\big(\mathcal{G}^{(l)}\big), \quad l = 0, \dots, L-1,
\end{equation}
where the initial graph structure $\mathcal{G}^{(0)}$ is fully determined by the template mesh $\mathcal{M}_t = \{\mathcal{V}_t, \mathcal{F} \}$ under our industry-standard topology setting, and we perform $L=4$ successive downsampling operations. Notably, we set the number of learned queries of Perceiver Resampler $N$ to match the number of vertices in the coarsest, i.e., the bottom-level graph $\mathcal{G}^L$, such that each query serves as a semantic anchor aligned with a vertex in $\mathcal{G}^L$. Since all meshes share a consistent topology, the pooling mapping is constructed once on the template mesh $\mathcal{M}_t$ and reused across all instances. Graph unpooling reverses this cached mapping in the decoder, and a final MLP head translates vertex features on $\mathcal{G}^{(0)}$ to 3D coordinates:
\begin{align}
    \hat{\mathcal{V}} &= \mathcal{D}(\mathcal{L}), \quad
    \hat{\mathcal{M}} = (\hat{\mathcal{V}}, \mathcal{F}).
\end{align}
where $\hat{\mathcal{V}}$ and $\hat{\mathcal{M}}$ are output vertices and head meshes, respectively.

\subsubsection{TOPOS-VAE Training}
Our TOPOS-VAE is trained on the head mesh dataset under the consistent topology. Inspired by recent work~\cite{wang2026high}, the objective combines a basic \textbf{vertex loss $L_{\text{vertex}}$}, which is defined as the $L_2$ distance between the decoded vertices $\hat{\mathcal{V}}$ and ground-truth $\mathcal{V}$, with three additional geometric losses that enforce a smooth surface and regular face connectivity.

\textbf{Face normal consistency loss $L_{\text{normal}}$} penalizes angular deviation between decoded and ground-truth face normals:
\begin{equation}
    L_{\text{normal}} = \frac{1}{N_f}\sum_{f \in \mathcal{F}} \left(1 - \frac{\hat{\mathbf{n}}_f \cdot \mathbf{n}_f}{|\hat{\mathbf{n}}_f||\mathbf{n}_f|}\right).
\end{equation}
For each triangle $f \in \mathcal{F}$, we compute the face normal $\mathbf{n}_f$ from its edge vectors via the cross product and $l_2$ normalization.

\textbf{Face angle consistency loss $L_{\theta}$} constrains local triangle shape to prevent degeneration by supervising interior cosines. For triangle $f$ with vertices $v_0, v_1, v_2$, $c_f^k$ denotes the cosine of the interior angle at $v_k$, which is computed from the edge vectors originating from $v_k$. $L_{\theta}$ is then defined as the $L_1$ distance between the decoded and ground-truth cosine values:
\vspace{-1mm}
\begin{equation}
L_{\theta} = \frac{1}{N_f}\sum_{f \in \mathcal{F}} \Big(\left|\hat{c}_f^0 - c_f^0\right| + \left|\hat{c}_f^1 - c_f^1\right|\Big).
\end{equation}

\vspace{-1mm}
\textbf{Discrete Gaussian curvature loss $L_{\text{gc}}$} preserves fine surface curvature via the signed dihedral angle on each interior edge $e$ shared by two faces with normals $\mathbf{n}_a, \mathbf{n}_b$:
\vspace{-1mm}
\begin{equation}
\kappa_e = \frac{1}{\pi} \cdot \text{sgn}\left((\mathbf{n}_a \times \mathbf{n}_b) \cdot \hat{\mathbf{e}}\right) \cdot \arctan\frac{|\mathbf{n}_a \times \mathbf{n}_b|}{\mathbf{n}_a \cdot \mathbf{n}_b},
\end{equation}
where $\hat{\mathbf{e}}$ is the unit edge direction and $\text{sgn}(\cdot)$ is the sign function encoding local convexity or concavity. $L_{\text{gc}}$ is defined as the $L_1$ distance between the decoded and ground-truth signed dihedral angle:
\vspace{-1mm}
\begin{equation}
L_{\text{gc}} = \frac{1}{N_e}\sum_e\left|\hat{\kappa}_e - \kappa_e\right|.
\end{equation}
\vspace{-1mm}

Finally, a \textbf{KL-divergence loss $L_{\text{KL}}$}~\cite{kingma2013auto} is adopted to regularize the distribution of latent token $\mathcal{L}$ toward a standard Gaussian. In summary, the total loss function of our TOPOS-VAE is defined as the weighted sum of all above losses:
\vspace{-1mm}
\begin{equation}
    L_{\text{VAE}} = \lambda_{\text{vertex}}L_{\text{vertex}} + \lambda_{\text{normal}} L_{\text{normal}} + \lambda_{\theta} L_{\theta} + \lambda_{\text{gc}} L_{\text{gc}} + \lambda_{\text{KL}} L_{\text{KL}}.
\end{equation}
\vspace{-1mm}

\subsection{TOPOS-DiT}
We employ rectified flow models~\cite{lipman2022flow} to generate the latent tokens $\mathcal{L}$ of TOPOS-VAE conditioned on the portrait images to realize our head mesh generation task. Rectified flow models use a linear interpolation forward process, $x(t) = (1-t)x_0 + t\epsilon$, which interpolates between data samples $x_0$ and noises $\epsilon$ with a timestep $t$. The backward process is represented as a time-dependent vector field, $v(x,t) = \nabla_t x$, moving noisy samples toward the data distribution. In our setting, we utilize a simple transformer backbone TOPOS-DiT to generate latent tokens $\mathcal{L} \in \mathbb{R}^{N\times d}$. The input noisy latent tokens, combined with positional embeddings, are fed into the TOPOS-DiT for denoising. Timestep information is incorporated using adaptive layer normalization (AdaLN) and a gating mechanism~\cite{peebles2023scalable}. For input images, we adopt visual features from DINOv2~\cite{oquab2023dinov2} and visual features are injected through cross attention layers as keys and values. The denoised latent tokens are further decoded into head meshes under the fixed reference topology.

\subsection{TOPOS-Texture}

\begin{figure}[t]
    \centering
    \includegraphics[width=\linewidth]{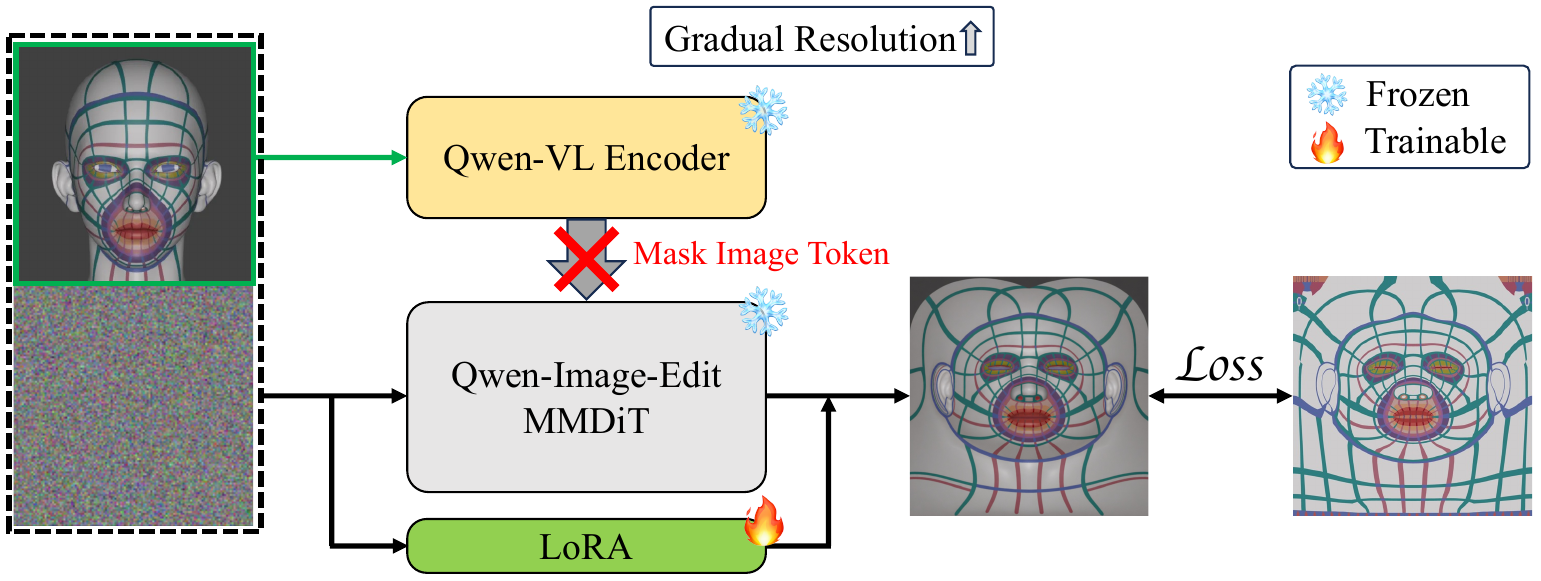}
    \vspace{-7mm}
    \caption{The pipeline of our proposed TOPOS-Texture.}
    \label{fig:texture-pipe}
    \vspace{-3mm}
\end{figure}

Inspired by~\cite{zeng2025renderformer}, we posit that generating an unwarpping texture map from a reference image is inherently learnable. However, high-quality facial textures are scarce and costly to acquire~\cite{bai2023ffhq}, making direct supervision challenging. We address this by leveraging a pretrained multimodal generative model, Qwen-Image-Edit~\cite{wu2025qwenimagetechnicalreport}, as a prior, reducing the learning task to UV-space unwrapping rather than texture synthesis from scratch. The pipeline of our TOPOS-Texture is shown in~\figref{fig:texture-pipe}.

To preserve the base model's generative capability under limited data, we apply LoRA~\cite{hu2021loralowrankadaptationlarge} to the Transformer~\cite{vaswani2017attention}'s attention layers, keeping the base weights, VAE and text encoder frozen. Recent multimodal generative models process reference images through two pathways: a self-attention stream governing texture synthesis, and a cross-attention stream conditioned on a vision-language (VL) encoder providing semantic control. Since VL encoders are trained on generic image-text data and lack UV-space awareness, feeding the reference images through the VL branch introduces conflicts between semantic guidance and desired spatial alignment. We therefore discard the image tokens from the VL branch, retaining only text hidden states as condition. 

We finetune TOPOS-Texture with the original Qwen-Image-Edit flow matching objective and a fixed text prompt adapted from the official template. Naively training on a small texture corpus at full resolution lets high-frequency detail dominate the loss, yielding textures that are locally sharp but globally misaligned in UV space. We therefore adopt a gradual resolution schedule: training begins at low resolution to establish UV alignment, and progressively introduces higher resolutions to recover sharpness. To keep the noise schedule well-calibrated across stages, we further apply a dynamic time-shift $\mu$ to the logit-normal noise sampling distribution, adapting $\mu$ to the image sequence length at each resolution.  Together, the resolution schedule and adaptive noise schedule reliably decouple the learning of UV alignment from the recovery of texture sharpness.

Furthermore, to obtain the texture map suitable for downstream rendering, the model is required to learn to disentangle intrinsic surface appearance from illumination. Inspired by~\cite{chen2024intrinsicanythinglearningdiffusionpriors, liang2025diffusionrendererneuralinverseforward}, we render training images under randomized lighting conditions, exposing the model to diverse shading variations while supervising against the same canonical texture map. This encourages the model to implicitly disentangle albedo from lighting, producing texture maps that are free of baked-in illumination.
\section{Experiments}

\begin{figure}[t]
    \centering
    \includegraphics[width=\linewidth]{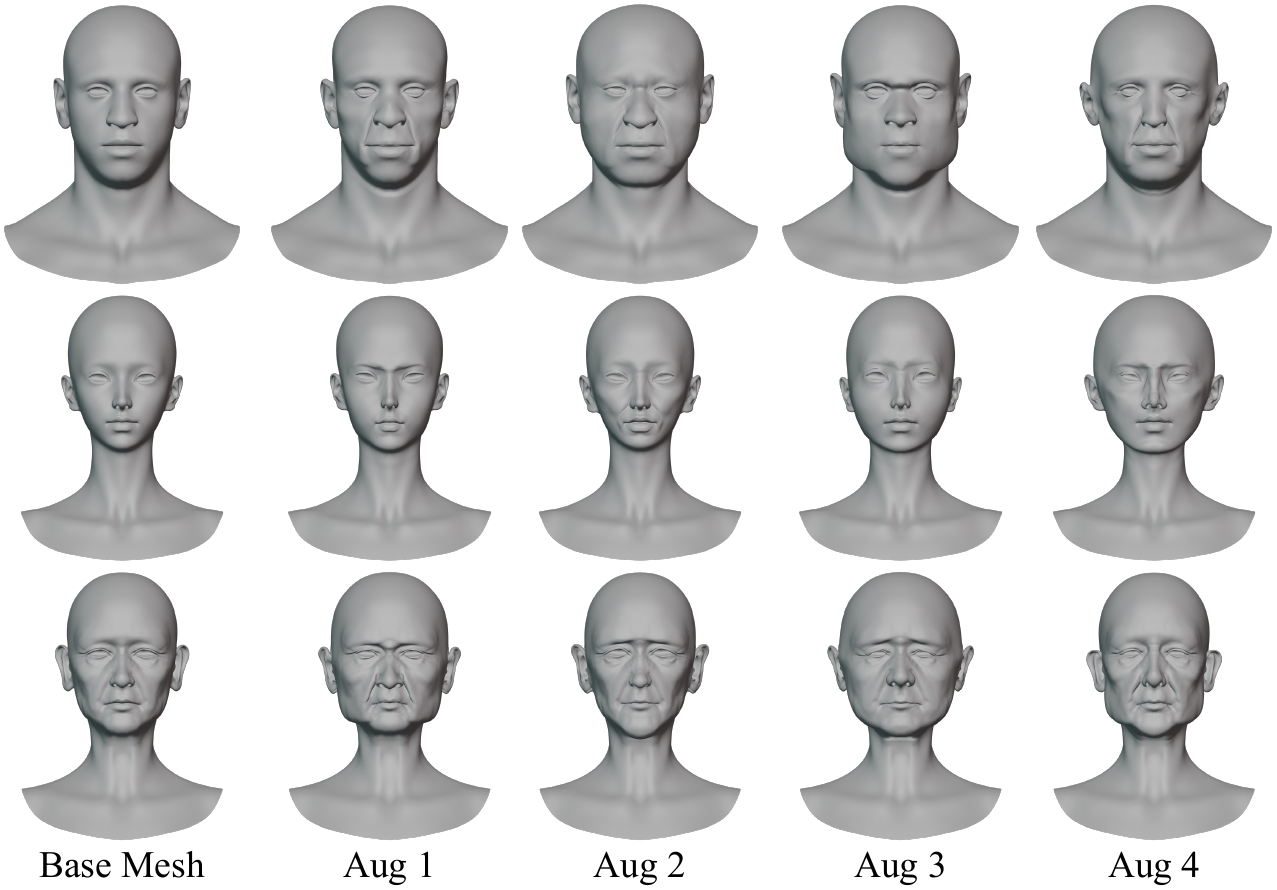}
    \vspace{-7mm}
      \caption{The results of our designed geometry augmentation algorithm. Aug 1--4 denote four independent variants augmented from the same base mesh. Please zoom in for better inspection.}
    \label{fig:geo-aug}
    \vspace{-3mm}
\end{figure}

\begin{figure}[t]
    \centering
    \includegraphics[width=\linewidth]{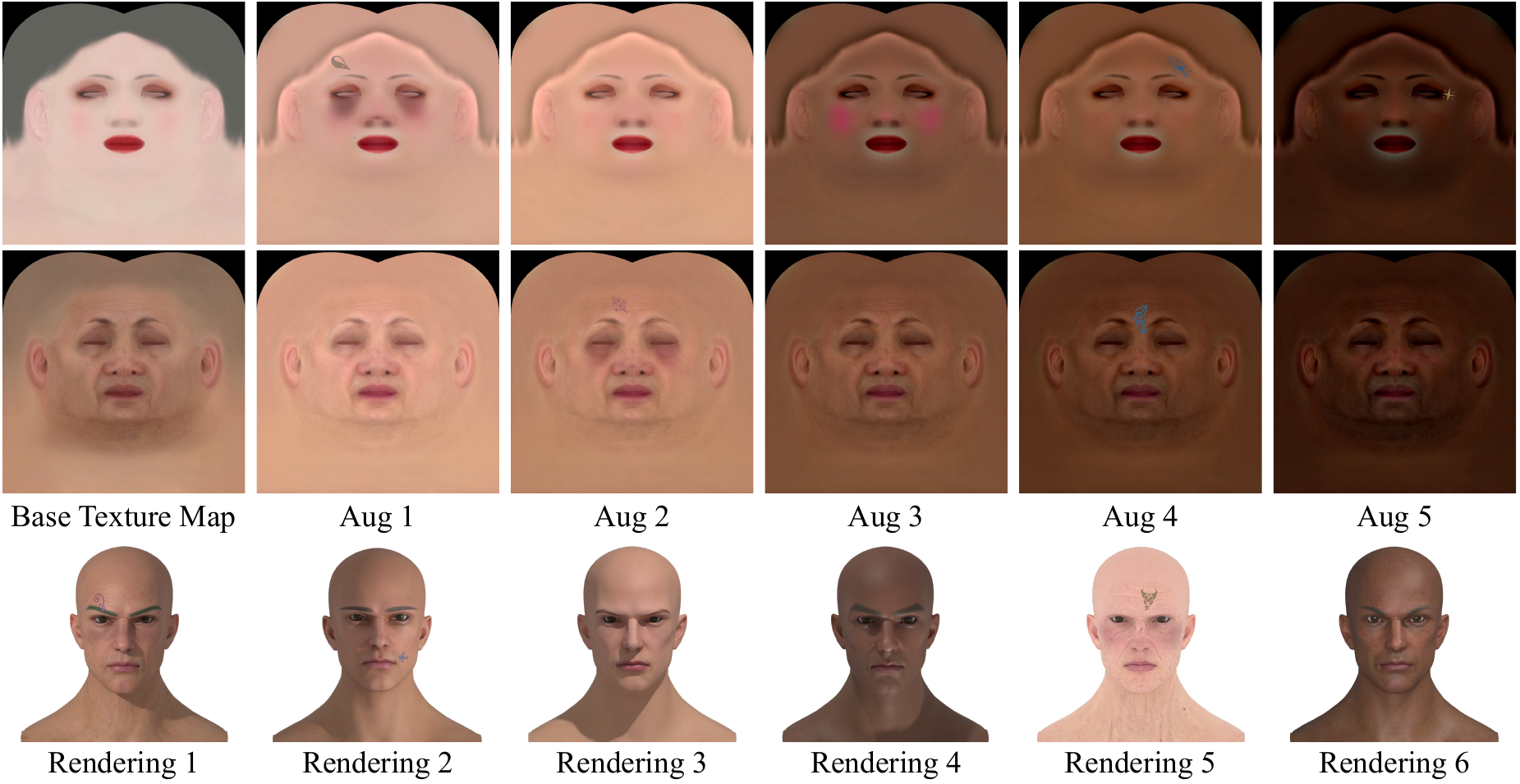}
    \vspace{-7mm}
      \caption{The first two lines show the results of our texture map augmentation. Aug 1--5 denote five independent variants augmented from the same base texture map. The last row shows six rendering images of the same head mesh using different texture maps.}
  \label{fig:texture-aug}
    \vspace{-2mm}
\end{figure}

\begin{figure*}[t!]
    \includegraphics[width=\textwidth]{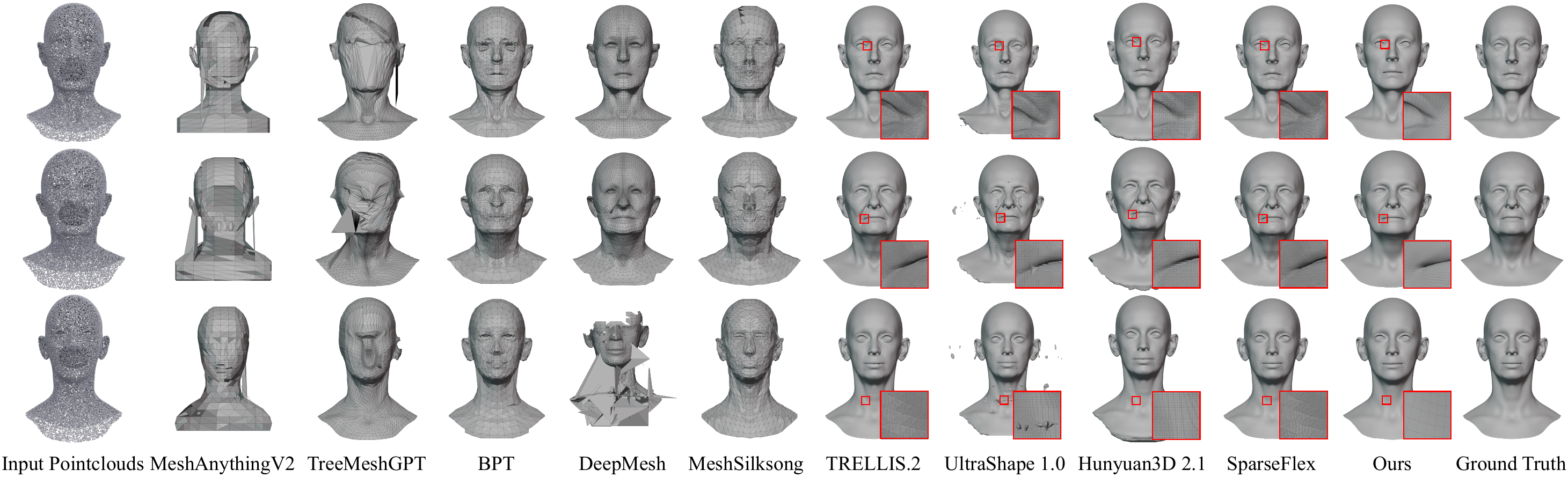}
    \vspace{-7mm}
    \caption{Qualitative results reconstructed by different methods on our test dataset $D_\text{test}$. The insets in the red boxes illustrate the topology and connectivity of the reconstructed head meshes. Please zoom in for better inspection.}
    \label{fig:vae-compare}
\end{figure*}

\begin{figure}[t]
    \centering
    \includegraphics[width=\linewidth]{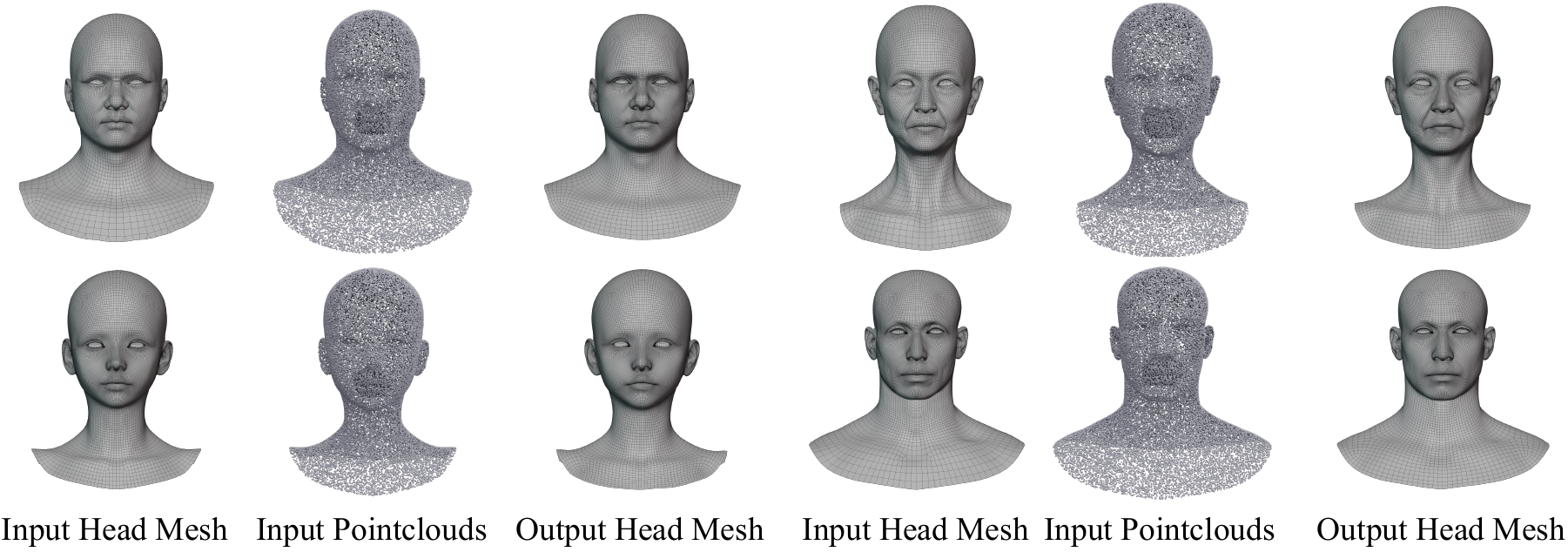}
    \vspace{-7mm}
      \caption{The first and second rows show conversion results from two different head mesh topologies. Please zoom in for better inspection.}
  \label{fig:topology}
\end{figure}

\subsection{Data Preparation}
We adopt the MetaHuman~\cite{EpicGames2021Metahuman} head mesh, which contains 24,049 vertices, 48,050 edges and 24,002 faces as our uniform reference topology. Our base dataset $D_0$ consists of 388 in-house head meshes, with 226 sharing the MetaHuman topology collected from internal 3D animation and video game productions, plus 162 in non-MetaHuman topologies drawn from our internal asset library.

Building upon this, we design a geometric augmentation algorithm to increase data diversity. Specifically, we derive 500 augmented variants from each of the 388 base head meshes in $D_0$, resulting in the final dataset $D$, which contains $388 \times 500 = 194{,}000$ head meshes in total. Among these, we denote the subset of head meshes with MetaHuman topology as $D_M$, comprising $226 \times 500 = 113{,}000$ head meshes in total. ~\figref{fig:geo-aug} visualizes the results of our proposed geometric augmentation, with algorithm details provided in the supplementary material.

For appearance and rendering, we first collect 356 different base texture maps from $D$. For each base texture map, we randomize 14  skin tones using the official MetaHuman implementation. Then, we apply procedural makeup augmentation over predefined semantic regions (e.g., eyelashes, blush, tattoos). We further perturb PBR material strengths and randomize lighting via combinations of area, point and ambient lights, with camera poses slightly perturbed around the frontal view. All meshes are rendered without hair to avoid occlusion. For each mesh, we render eight images using eight randomly chosen texture maps, and select one per training step of TOPOS-DiT and TOPOS-Texture. ~\figref{fig:texture-aug} shows the texture augmentation results and different render images of the same head mesh.

Our test set $D_\text{test}$ contains 269 high-quality textured 3D head meshes in MetaHuman topology derived from licensed digital human assets (including 66 official MetaHuman assets), covering diverse identities and appearances. The frontal-view rendering of each mesh serves as the input condition. The underlying textured meshes in $D_\text{test}$ are used solely for evaluation and are not redistributed.

\subsection{Training Details}
TOPOS-VAE requires consistent topology and is trained on $D_M$ only. Once trained, it can encode head meshes with diverse topologies. Therefore, TOPOS-DiT and TOPOS-Texture are both trained on the full $D$. All training runs on 16 NVIDIA H20 GPUs in PyTorch~\cite{paszke2019pytorch}, taking 3, 7 and 2 days for TOPOS-VAE, TOPOS-DiT and TOPOS-Texture, respectively. More details of experimental settings and hyperparameters are provided in the supplementary material.

\subsection{3D Head Mesh Reconstruction}
We evaluate TOPOS-VAE reconstruction on $D_\text{test}$ against two families of baselines: (i) direct mesh generation methods that auto-regressively emit artist-like meshes from pointclouds, including DeepMesh~\cite{zhao2025deepmesh}, TreeMeshGPT~\cite{lionar2025treemeshgpt}, MeshAnythingV2~\cite{chen2025meshanything}, BPT~\cite{weng2024scaling}, MeshSilksong~\cite{songtopology} and (ii) neural-field VAEs that recover dense meshes via post-processing algorithm, including Hunyuan3D 2.1~\cite{hunyuan3d2025hunyuan3d}, SparseFlex~\cite{he2025sparseflex}, TRELLIS.2~\cite{xiang2025trellis2}, UltraShape 1.0~\cite{jia2025ultrashape}. Following ConvONet~\cite{peng2020convolutional}, we report $L_1$ Chamfer distance (CD), normal consistency (NC) and F-Score against ground truth head meshes. The details of these metrics are provided in the supplementary material.

\paragraph{Qualitative Results}
~\figref{fig:vae-compare} compares reconstructions on $D_\text{test}$. Direct mesh generation methods emit low-poly head meshes but exhibit noticeable artifacts and distorted structures. Neural-field models produce smooth surfaces yet generate excessive vertices and struggle with non-manifold head meshes. In contrast, TOPOS-VAE faithfully reconstructs input pointclouds with a clean and consistent topology. Moreover, TOPOS-VAE can also act as a topology unifier. As shown in~\figref{fig:topology}, it is capable of converting pointclouds sampled from meshes with diverse topologies into the MetaHuman topology while preserving geometric details and identity.

\vspace{-2mm}
\paragraph{Quantitative Results}
TOPOS-VAE outperforms all baselines on all metrics at the top part of~\tabref{tab:vae}, while preserving exactly the input vertex and face counts. Neural-field reconstruction models produce millions of vertices, unsuitable for downstream tasks. Direct mesh generation methods require auto-regressive decoding of thousands of tokens that takes several minutes. TOPOS-VAE is non-autoregressive and post-processing free, enabling reconstruction within one second, markedly faster than all compared methods.

\begin{table}[t]
  \centering
  \caption{Quantitative results of 3D head meshes reconstruction on our test dataset $D_\text{test}$. The top of the table shows the comparison with other 3D reconstruction methods while the bottom presents our ablation studies and analysis. V-num and F-num denote the average number of vertices and faces of the reconstructed 3D head meshes of each method, respectively.}
  \vspace{-4mm}
  \resizebox{\columnwidth}{!}{
  \begin{tabular}{lcccccc}
      \toprule[1pt]
      {Method} & CD $\downarrow$ & NC $\uparrow$ &  F-Score $\uparrow$ & V-num & F-num \\
      \midrule[1pt]
       
       MeshAnythingV2~\cite{chen2025meshanything}  & 0.119 & 0.6546 & 0.0513 & 1236.78 & 2097.64 \\

       TreeMeshGPT~\cite{lionar2025treemeshgpt} & 0.0864 & 0.818 & 0.0751 & 4341.90 & 8073.21 \\

       BPT~\cite{weng2024scaling} & 0.0417 & 0.900 & 0.113 & 1518.92 & 3001.90 \\

       DeepMesh~\cite{zhao2025deepmesh} & 0.131 & 0.624 & 0.0513 & 8564.66 & 16225.93 \\

       MeshSilksong~\cite{songtopology} & 0.0641 & 0.867 & 0.0678 & 2364.97 & 4659.99 \\

       TRELLIS.2 (512)~\cite{xiang2025trellis2} & 0.124 & 0.787 & 0.0309 & 508577.93 & 1018609.32 \\

       TRELLIS.2 (1024)~\cite{xiang2025trellis2} & 0.124 & 0.787 & 0.0309 & 2032858.15 & 4070818.52 \\

       UltraShape 1.0~\cite{jia2025ultrashape} & 0.119 & 0.605 & 0.0598 & 1346200.94 & 2689415.33 \\

       Hunyuan3D 2.1~\cite{hunyuan3d2025hunyuan3d} & 0.151 & 0.664 & 0.0330 & 634685.12 & 1269357.18 \\

       SparseFlex (512)~\cite{he2025sparseflex} & 0.124 & 0.786 & 0.0309 & 514463.00 & 1027306.41 \\

       SparseFlex (1024)~\cite{he2025sparseflex} & 0.124 & 0.787 & 0.0310 & 2047892.75 & 4092383.40 \\
     
    \midrule[1pt]
    
    VecSet-Learn~\cite{zhang20233dshape2vecset} & 0.0132 & \underline{0.967} & 0.502 & 24049 & 24002 \\
    
    VecSet-FPS~\cite{zhang20233dshape2vecset} & 0.0439 & 0.833 & 0.156 & 24049 & 24002 \\
    
    w/o Geo Loss & \underline{0.0094} & 0.93 & \underline{0.636} & 24049 & 24002 \\
    
    TOPOS-VAE & \textbf{0.0055} & \textbf{0.987} & \textbf{0.915} & 24049 & 24002 \\
\bottomrule[1pt]
  \end{tabular}
  }
  \vspace{-2mm}
  \label{tab:vae}
\end{table}

\begin{figure}[t]
    \centering
    \includegraphics[width=\linewidth]{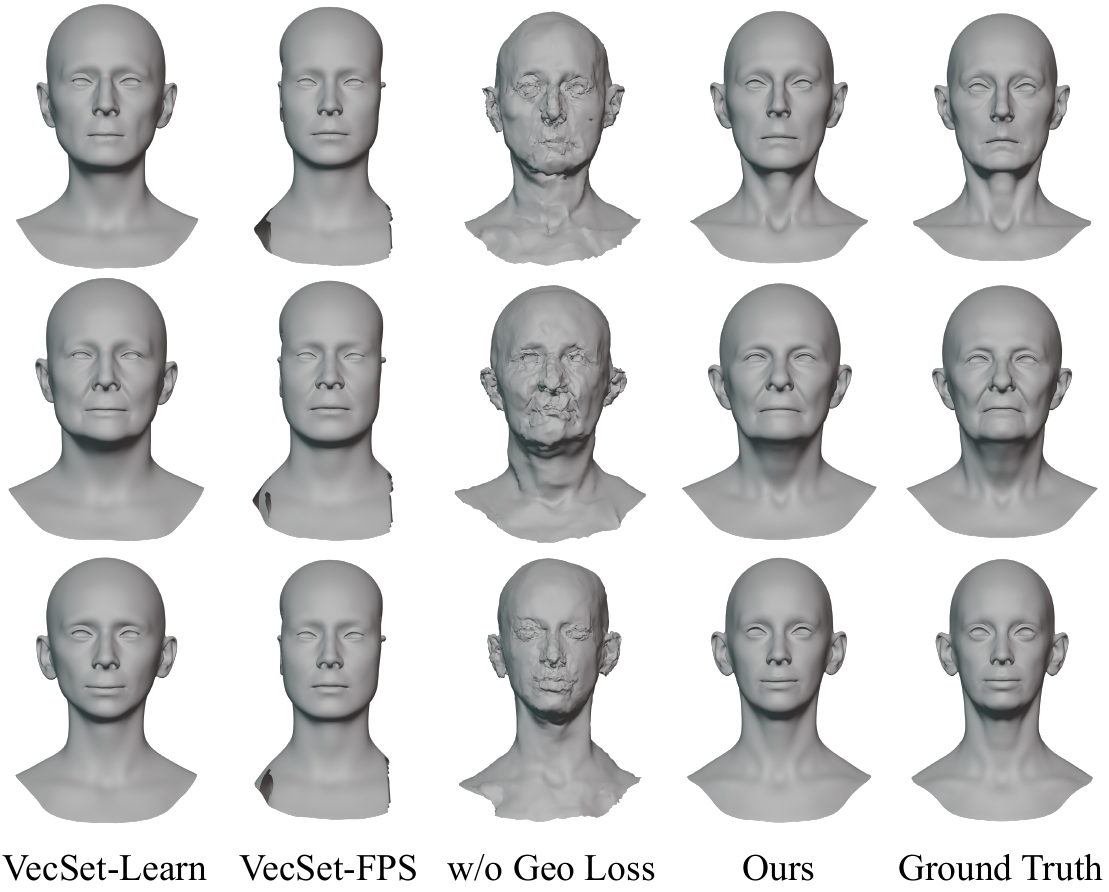}
    \vspace{-7mm}
      \caption{Qualitative results of our ablation studies on TOPOS-VAE using the same three head meshes in~\figref{fig:vae-compare}.}
  \label{fig:ablation}
\end{figure}

\begin{figure}[t]
    \centering
    \includegraphics[width=\linewidth]{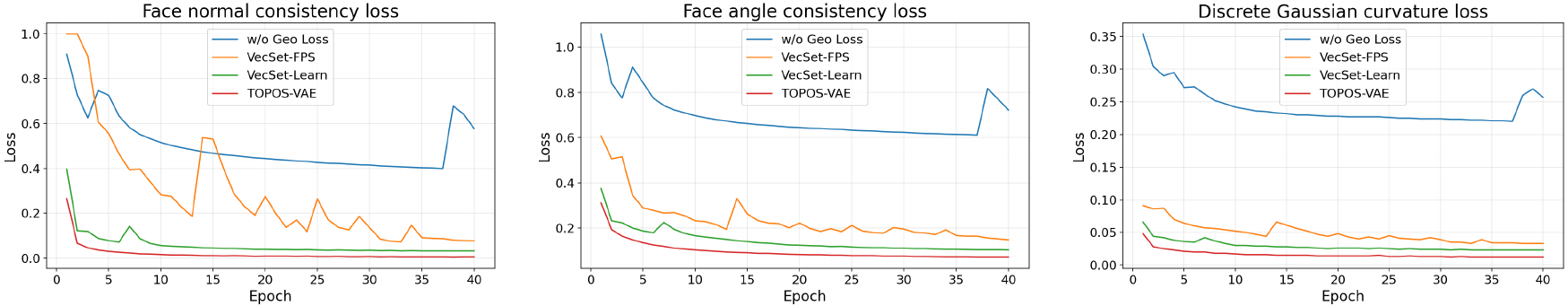}
    \vspace{-7mm}
      \caption{Training loss curves of three geometry losses of different variants. Please zoom in for better inspection.}
  \label{fig:loss-curve}
  \vspace{-3mm}
\end{figure}

\vspace{-2mm}
\paragraph{Ablation Studies}
We ablate two design choices in~\figref{fig:ablation} and the bottom part of~\tabref{tab:vae}. We also visualize the training loss curves of three geometry losses in~\figref{fig:loss-curve}. First, we replace the Perceiver Resampler with VecSet~\cite{zhang20233dshape2vecset} encoders using learnable queries (VecSet-Learn) or FPS queries (VecSet-FPS). The former converges slowly and loses fine facial details, while the latter fails to reconstruct complete head meshes. Second, training only with $L_\text{vertex}$ (``w/o Geo Loss'') yields significant structural artifacts and irregular topology, confirming the necessity of geometric supervision.

\subsection{3D Head Mesh Generation}
We evaluate TOPOS-DiT on our test set $D_\text{test}$, using the frontal view rendered images as input. We compare it with several 3DMM-based face reconstruction methods, including FFHQ-UV~\cite{bai2023ffhq}, FLAME 2020~\cite{FLAME2017}, which is implemented in FreeUV~\cite{yang2025freeuv}, UV-IDM~\cite{li2024uv}, HRN~\cite{lei2023hierarchical} and Pixel3DMM~\cite{giebenhain2025pixel3dmm}, as well as representative 3D shape generation models, including DreamFace~\cite{zhang2023dreamface}, Hunyuan3D 2.1~\cite{hunyuan3d2025hunyuan3d}, TRELLIS.2~\cite{xiang2025trellis2}, UltraShape 1.0~\cite{jia2025ultrashape} and commercial models Model A and Model B. All these methods accept single portrait image as input and output 3D head mesh. For closed-source DreamFace and two commercial models, we report only qualitative comparisons through their official web interfaces.

\begin{table}[t]
  \centering
  \caption{Quantitative results of 3D head meshes generation on our test dataset $D_\text{test}$. V-num and F-num denote the average number of vertices and faces of the generated 3D head meshes of each method, respectively.}
  \vspace{-3mm}
  \resizebox{\columnwidth}{!}{
  \begin{tabular}{lcccccc}
      \toprule[1pt]
      {Method} & CD $\downarrow$ & NC $\uparrow$ &  F-Score $\uparrow$ & V-num & F-num \\
      \midrule[1pt]
       
       FFHQ-UV~\cite{bai2023ffhq}  & \underline{0.0150} & \underline{0.940} & 0.564 & 20792 & 40832 \\
       
       FLAME 2020~\cite{FLAME2017} & 0.0217 & 0.906 & 0.458 & 5118 & 9976 \\
       
       UV-IDM~\cite{li2024uv} & 0.0461 & 0.812 & 0.438 & 53215 & 105954 \\
       
       HRN~\cite{lei2023hierarchical} & 0.0634 & 0.777 & 0.342 & 35709 & 70789 \\
       Pixel3DMM~\cite{giebenhain2025pixel3dmm} & 0.0185 & 0.920 & 0.515 & 5023 & 9976 \\
       
       Hunyuan3D 2.1~\cite{hunyuan3d2025hunyuan3d} & 0.0218 & 0.870 & \underline{0.667} & 746179.62 & 1492399.77 \\
       
       TRELLIS.2 (1024)~\cite{xiang2025trellis2} & 0.0349 & 0.808 & 0.520 & 329620.02 & 485600.25 \\
       
       UltraShape 1.0~\cite{jia2025ultrashape} & 0.0225 & 0.866 & 0.659 & 5147079.16 & 10294336.87 \\
       
       Ours & \textbf{0.00613} & \textbf{0.978} & \textbf{0.857} & 24049 & 24002 \\
       
\bottomrule[1pt]
  \end{tabular}
  }
  \vspace{-2mm}
  \label{tab:dit}
\end{table}

\begin{figure*}[t!]
\vspace{-2mm}
    \includegraphics[width=\textwidth]{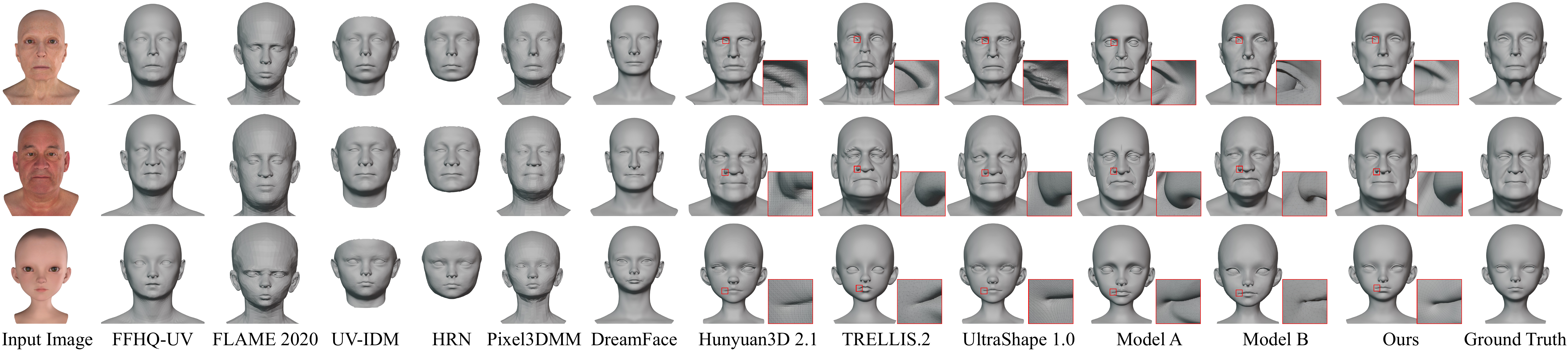}
    \vspace{-7mm}
    \caption{Qualitative results generated by different methods on our test dataset $D_\text{test}$. The insets in the red boxes illustrate the topology and connectivity of the generated head meshes. Please zoom in for better inspection.}
    \label{fig:geo-compare}
\end{figure*}

Different methods produce meshes in heterogeneous coordinate frames, scales and spatial extents. Some only generate the frontal face shell, while ground truth covers the full head mesh down to the neck and shoulders. Therefore, we align each generated mesh $\mathcal{M}_G$ to its paired ground truth $\mathcal{M}_R$ via a 7-DoF similarity transform solved by trimmed similarity ICP~\cite{besl1992method, zinsser2005point, chetverikov2005robust} with multi-start initialization, where trimming restricts the fit to mutually visible facial regions. We then clip $\mathcal{M}_G$ to the bounding box of $\mathcal{M}_R$ inflated by 5\% (via plane-based triangle slicing) while leaving $\mathcal{M}_R$ intact, so incomplete predictions are still penalized. We report CD, NC and F-Score with all distances normalized by the bounding-box diagonal of $\mathcal{M}_R$. Full implementation details of 3D head mesh alignment algorithm are provided in the supplementary material.
  
\paragraph{Qualitative Results}
~\figref{fig:geo-compare} compares our method with the competing approaches. 3DMM-based methods recover plausible global shapes but miss fine geometric detail due to the limited expressiveness of parameters. General 3D generative models produce meshes with excessive vertices and irregular topology, as highlighted in red boxes. In contrast, TOPOS-DiT produces head meshes with clean topology, smooth connectivity and faithful identity fidelity, striking a balance between geometric expressiveness and mesh regularity.

\vspace{-2mm}
\paragraph{Quantitative Results}
\tabref{tab:dit} demonstrates that TOPOS-DiT outperforms all baselines across CD, NC and F-Score. The substantially higher F-Score over 3DMM-based methods reflects our finer geometric detail beyond the parametric subspace. Compared with general 3D generative models, which produce millions of irregular polygons, our generative results remain compact, well-structured and directly usable in downstream pipelines.

\begin{figure}[t]
    \centering
    \includegraphics[width=\linewidth]{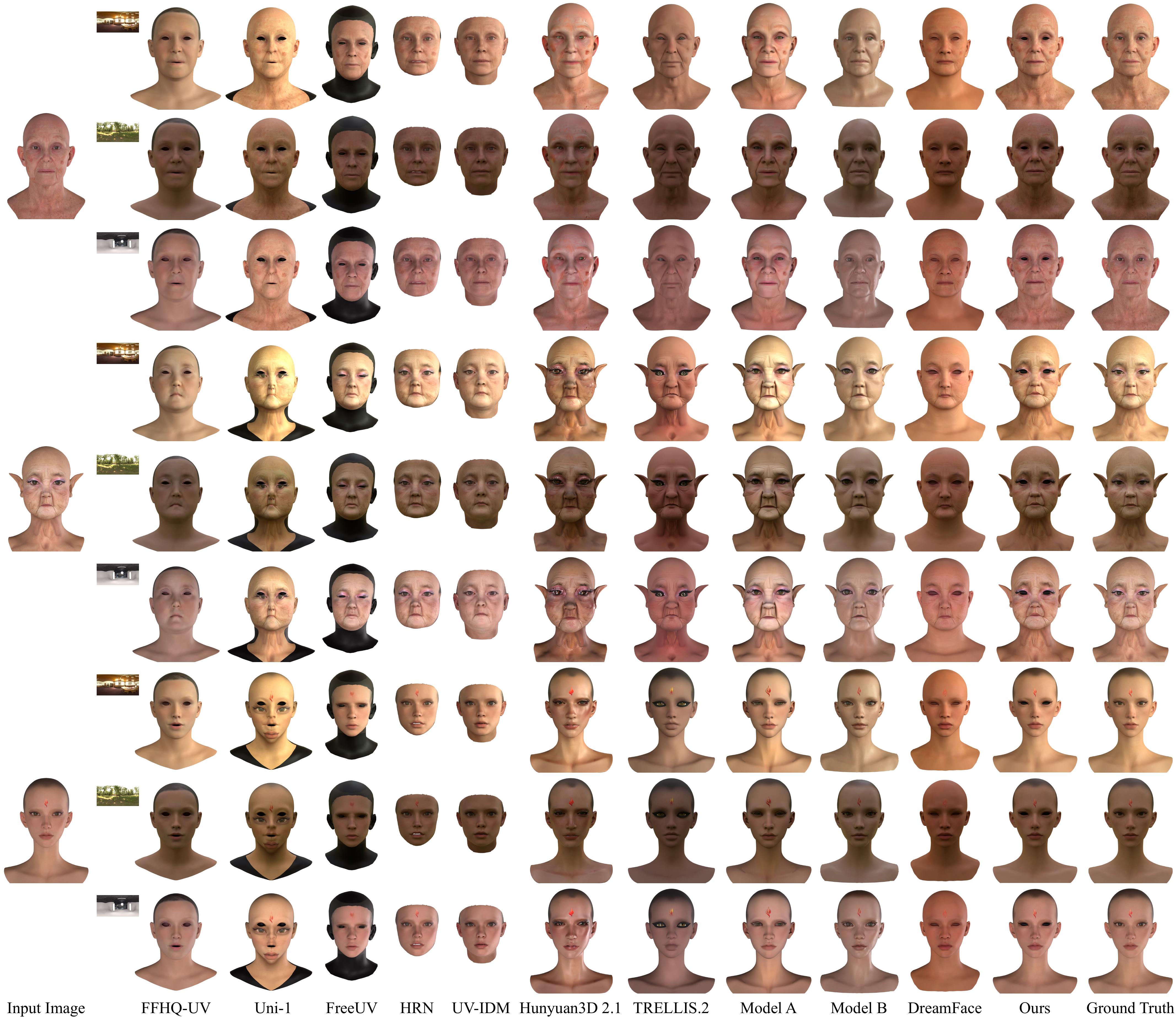}
    \vspace{-7mm}
      \caption{Qualitative results generated by different methods on our test dataset $D_\text{test}$. The upper-left insets illustrate the environment maps. Please zoom in for better inspection.}
  \label{fig:render}
  \vspace{-1mm}
\end{figure}

\subsection{3D Head Texture Generation}
We compare TOPOS-Texture with methods capable of generating texture maps introduced in the previous subsection, the same two commercial models, as well as the recent multimodal reasoning model Uni-1~\cite{luma_uni1}, which produces a single unwrapped UV map from three multi-view photographs of the same subject.

Beyond $D_\text{test}$, we further evaluate on 159 uncurated in-the-wild portrait images (82 males and 77 females) collected from public web sources, academic datasets~\cite{karras2019style, liu2015faceattributes} and online AI image generation tools. Since our framework expects near-frontal and unobstructed portraits to match the training rendering strategy, we adopt a fine-tuned FLUX.1 Kontext~\cite{labs2025flux1kontextflowmatching} model to frontalize the face and remove major occluders of input images to satisfy this requirement and the resulting edited images serve as the actual inputs fed into all compared methods, which has become a common practice in recent work~\cite{wang2026high}. It is worth noting that Uni-1~\cite{luma_uni1} uses the head mesh of FFHQ-UV~\cite{bai2023ffhq} and we only conduct qualitative comparisons for it on $D_\text{test}$ using three multiview renderings of the ground truth mesh, along with the official prompts provided on its website.

\begin{figure*}[t!]
    \includegraphics[width=\textwidth]{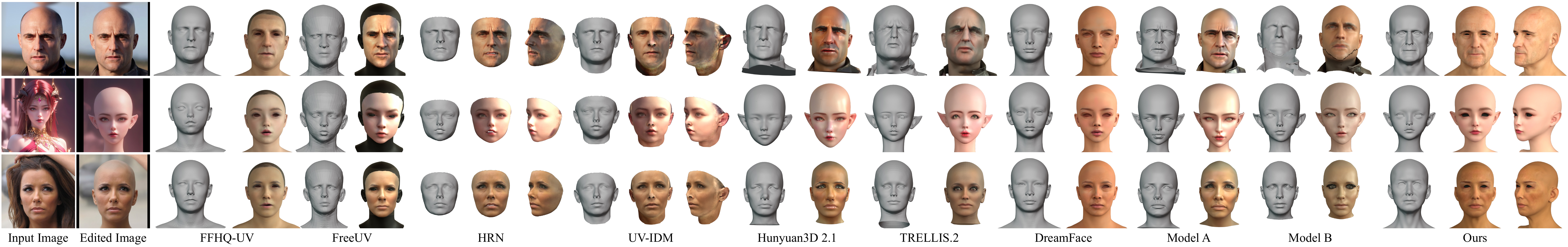}
    \vspace{-7mm}
    \caption{Qualitative results generated by different methods on in-the-wild images. For each method, we show both the geometry and texture renderings under the same environment map side by side. For HRN, UV-IDM and our method, we additionally include a profile view rendering. Please zoom in for better inspection.}
    \label{fig:in-the-wild}
\end{figure*}

On $D_\text{test}$, we render generated and ground-truth meshes from seven viewpoints (one frontal + six random within the frontal hemisphere) under six lighting environments at four illumination angles each. For each identity, we apply a facial mask, which is derived from the ground-truth mesh to exclude eyes, neck and shoulders so that extraneous geometry does not dominate the metrics. We report LPIPS~\cite{zhang2018unreasonable}, FID~\cite{parmar2022aliased}, KID~\cite{binkowski2018demystifying} and Cosine SIMilarity of identity features (CSIM)~\cite{deng2019arcface}. On in-the-wild images, since no ground-truth meshes are available, we align all competing meshes to ours using the same alignment algorithm in previous subsection. Then, we render from the frontal view under the same lighting settings and compute FID, KID as well as CSIM against the input edited image. To measure the geometric similarity, we additionally render an untextured version of each mesh and report CSIM* against the input edited image. To remove the influence of eyes generated by particular methods, we mask the eye region in all renderings using convex hulls of landmarks detected by LivePortrait~\cite{guo2024liveportrait}. The detailed of these metrics are provided in the supplementary material.

\paragraph{Qualitative Results.}
~\figref{fig:render} shows qualitative comparisons on $D_\text{test}$, where each three-row group renders one identity under three environment maps. Our method consistently produces high-fidelity skin textures and accurate color reproduction across all lighting conditions, closely matching the ground truth. 3DMM-based methods and DreamFace fail to recover out-of-distribution geometry such as elongated ears. FFHQ-UV, TRELLIS.2 and two commercial models produce overly smooth, flat textures lacking fine skin detail. Uni-1 exhibits catastrophic texture-geometry misalignment and Hunyuan3D 2.1 shows noisy color artifacts with baked-in shading that compromise relightability.

In-the-wild comparisons are shown in~\figref{fig:in-the-wild}, where each method displays geometry and texture renderings under the same environment map. In terms of geometry, 3DMM-prior methods and DreamFace are limited by their shape bases (e.g., failing on the elongated ears in row 2), while our method recovers identity-specific details such as nasolabial folds and chin structure. For texture, baked-in illumination is a common failure mode in HRN, UV-IDM, Hunyuan3D 2.1 and Model A, whereas our method recovers clean albedo that decouples appearance from lighting and remains faithful to each subject's intrinsic skin tone across novel views.

\vspace{-3mm}
\paragraph{Quantitative Results.}
The top part of~\tabref{tab:texture_results} reports the comparison on both $D_\text{test}$ and in-the-wild images. On $D_\text{test}$, TOPOS-Texture is the best across all four metrics, with particularly large margins on FID and KID. On in-the-wild images, our method again attains the lowest FID and KID. The CSIM gap to HRN and UV-IDM stems from those methods baking the original illumination into their textures, which inflates identity similarity under the input viewpoint but degrades under novel views, as demonstrated in the profile view renderings in~\figref{fig:in-the-wild}. CSIM* further reflects this effect by evaluating untextured renderings, where our method achieves the highest score.

\begin{table}[t]
  \centering
  \caption{Quantitative results of 3D head texture generation on test dataset $D_\text{test}$ and in-the-wild images. The top of the table shows the comparison with other methods while the bottom presents our ablation studies.}
  \vspace{-3mm}
  \resizebox{\linewidth}{!}{
  \begin{tabular}{lccccccccc}
      \toprule[1pt]
      \multirow{2}[2]{*}{Method} & \multicolumn{4}{c}{$D_\text{test}$} & \multicolumn{4}{c}{In-the-wild} \\
      \cmidrule[0.5pt](lr){2-5} \cmidrule[0.5pt](lr){6-9}
       & LPIPS $\downarrow$ & FID $\downarrow$ & KID $\downarrow$ & CSIM $\uparrow$ & FID $\downarrow$ & KID $\downarrow$ & CSIM $\uparrow$ & CSIM* $\uparrow$ \\
      \midrule[1pt]
      FFHQ-UV~\cite{bai2023ffhq} & 0.0257 & 28.883 & 0.0334 & 0.738 & 58.093 & 0.0413 & 0.584 & 0.359 \\
      
      FreeUV~\cite{yang2025freeuv} & 0.0533 & 26.671 & 0.0253 & 0.526 & 98.967 & 0.0877 & 0.512 & 0.367 \\
      
      UV-IDM~\cite{li2024uv} & 0.0236 & 9.624 & 0.00664 & 0.752 & 63.276 & 0.0429 & \underline{0.694} & 0.416 \\
      
      HRN~\cite{lei2023hierarchical} & 0.0321 & 8.289 & 0.00554 & 0.773 & 91.572 & 0.0836 & \textbf{0.723} & 0.398 \\
      
      Hunyuan3D 2.1~\cite{hunyuan3d2025hunyuan3d} & 0.0319 & 20.009 & 0.0149 & 0.655 & 81.011 & 0.0352 & 0.551 & 0.394 \\
      
      TRELLIS.2~\cite{xiang2025trellis2} & 0.0875 & 57.233 & 0.0385 & 0.545 & 84.826 & 0.0346 & 0.521 & \underline{0.422} \\

      \midrule[1pt]

      w/o LoRA & \underline{0.0138} & \underline{2.678} & 0.00210 & 0.826 & 45.348 & 0.0221 & 0.611 & - \\
      
      w/ VL Img Tokens & 0.0142 & 2.992 & \underline{0.00203} & 0.789 & 43.683 & 0.0206 & 0.606 & - \\
      
      w/o Res. Sched. & 0.0139 & 2.933 & 0.00251 & \underline{0.852} & \underline{42.681} & \underline{0.0195} & 0.605 & - \\

      Ours & \textbf{0.0119} & \textbf{2.456} & \textbf{0.00199} & \textbf{0.855} & \textbf{41.066} & \textbf{0.0187} & 0.613 & \textbf{0.438} \\

      \bottomrule[1pt]
  \end{tabular}
  }
  \vspace{-2mm}
  \label{tab:texture_results}
\end{table}

\begin{figure}[t]
    \centering
    \includegraphics[width=\linewidth]{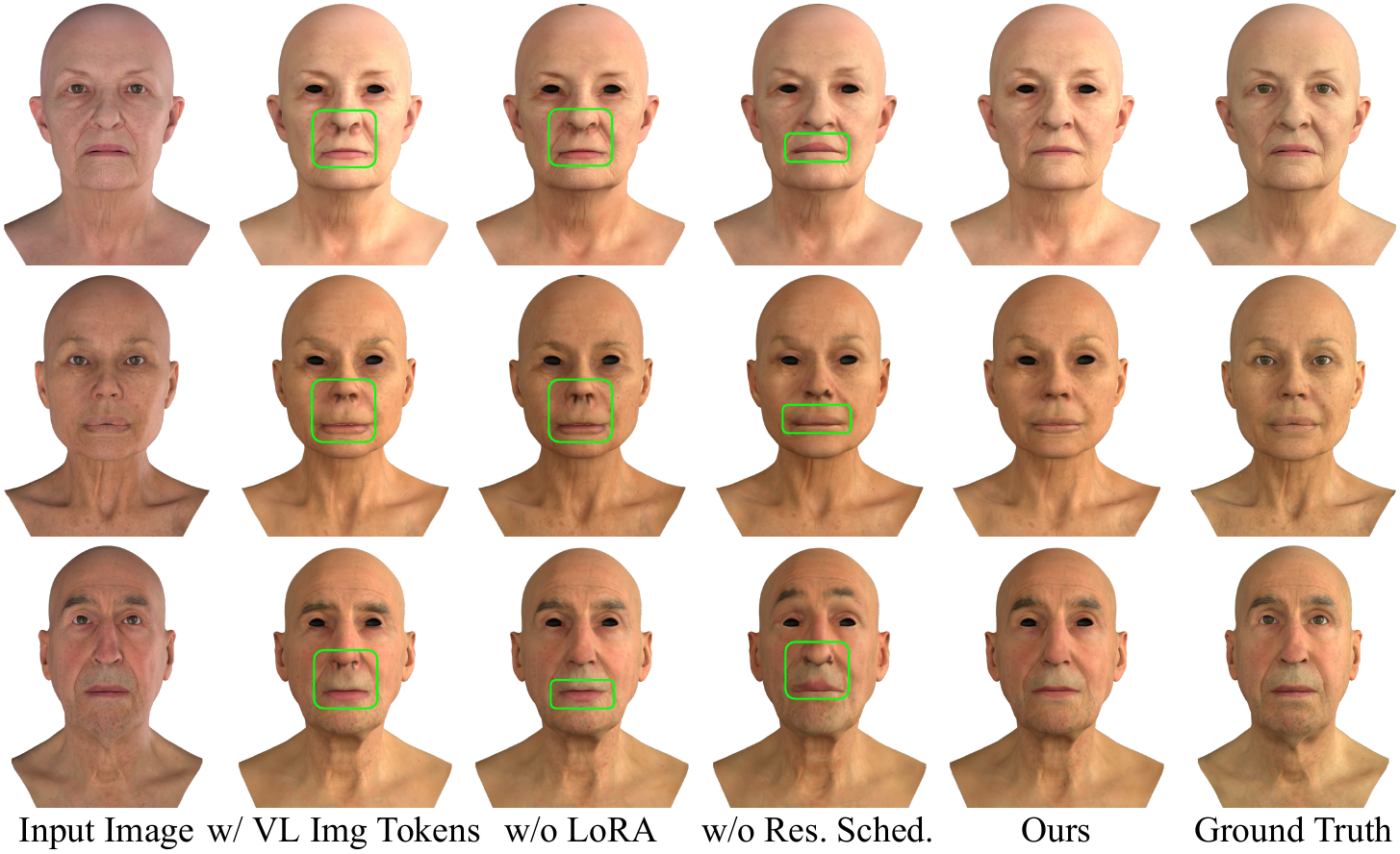}
    \vspace{-7mm}
      \caption{Qualitative results of our ablation studies on TOPOS-Texture. The green boxes highlight the texture-geometry misalignment.}
  \label{fig:texture-ablation}
  \vspace{-1mm}
\end{figure}

\vspace{-2mm}
\paragraph{Ablation Studies}
We validate our texture module's designs through three ablations: (1) retaining VL image tokens, (2) training directly at full resolution without the resolution schedule or adaptive time-shift $\mu$ and (3) unfreezing the backbone network (``w/o LoRA''). As reported in~\tabref{tab:texture_results}, all variants yield inferior metrics compared to our full method. We also show the visual effects of ablations in~\figref{fig:texture-ablation}, where these three variants exhibit notable texture-geometry misalignment, particularly in semantically sensitive regions such as nose and mouth. This consistent degradation confirms that discarding VL image tokens, the adaptive resolution schedule and frozen backbone are essential for high-quality texture unwrapping.

\subsection{Application}
\begin{figure*}[t!]
    \includegraphics[width=\textwidth]{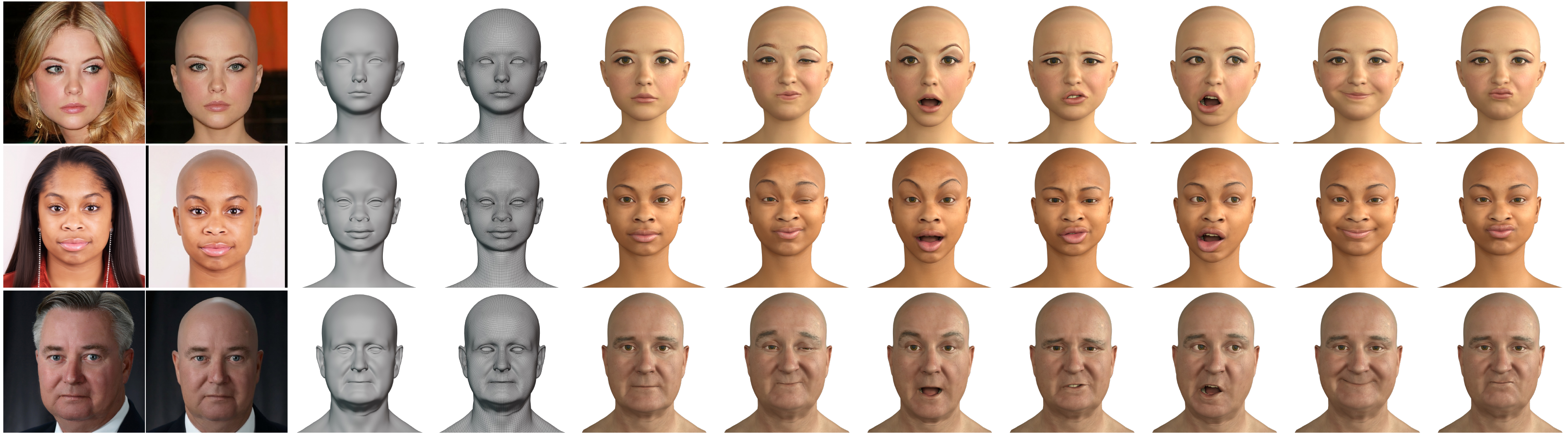}
    \vspace{-7mm}
    \caption{The animation results of our generated 3D head mesh. From left to right are the input image, edited image, generated geometry, mesh topology, generated texture and several animation results across different expressions, respectively.}
    \label{fig:animation}
\end{figure*}

Finally, we show that the industry-standard topology of our outputs seamlessly supports downstream production tasks. As shown in~\figref{fig:animation}, given a single portrait image, our TOPOS produces 3D head mesh that faithfully preserves geometry and appearance details. After bonding to the pre-defined facial rig, the generated meshes support rig-driven animation across diverse facial expressions, including mouth opening, eye blinking, brow furrowing and smiling, while maintaining geometric consistency and texture fidelity.

\section{Conclusion}
In this paper, we present TOPOS, a framework tailored for single image conditioned 3D head generation under an industry-standard, uniform topology. TOPOS combines a Perceiver Resampler with a GNN decoder for fixed-topology mesh modeling, a rectified flow transformer for 3D head geometry generation and a fine-tuned Qwen-Image-Edit model for relightable UV textures. Extensive experiments demonstrate state-of-the-art performance across 3D head mesh reconstruction, generation and texture generation, running orders of magnitude faster than manual artist workflows. Since TOPOS conforms to the studio-defined reference topology, its generated heads are directly compatible with production pipelines for downstream rigging, skinning and character animation. We hope TOPOS offers a scalable solution for digital human creation and inspires future work on generalizable 3D human digitization as well as uniform topology 3D mesh generation.

\bibliographystyle{ACM-Reference-Format}
\bibliography{bibliography}

\clearpage
\appendix
\section*{Table of Contents}
We first provide a brief overview of our supplementary material. This supplementary material consists of the following sections and
contents:

\begin{itemize}

\item \textbf{\secref{sec:hyperparameters}}: Experimental settings and hyperparameters details
\item \textbf{\secref{sec:augmentation}}: Geometric augmentation algorithm details.
\item \textbf{\secref{sec:alignment}}: Full implementation details of 3D head mesh alignment algorithm.
\item \textbf{\secref{sec:metrics}}: Evaluation metrics details.
\item \textbf{\secref{sec:results}}: Additional experimental results and visualizations.
\item \textbf{\secref{sec:ethics}}: Ethics statement of our method to avoid malicious use.

\end{itemize}

\section{Experimental Settings and Hyperparameters}
\label{sec:hyperparameters}
The encoder of our TOPOS-VAE uses 8 Perceiver Resampler blocks with hidden width 128, 8 attention heads of dimension 64 and an MLP expansion ratio of 4. The GNN decoder has $L = 4$ graph upsampling stages. The shape of latent tokens $\mathcal{L}$ is $1513 \times 32$. We optimize the network with AdamW optimizer~\cite{loshchilov2017decoupled} ($\beta_1 = 0.9$, $\beta_2 = 0.999$ and weight decay = 0.1). The initial learning rate is $1\times10^{-3}$, decayed with a polynomial schedule of power 0.9. The loss weights are set to $\lambda_\text{vertex} = 2$, $\lambda_\text{normal} = 1$, $\lambda_\theta = 5$, $\lambda_\text{gc} = 5$, and $\lambda_\text{KL} = 1\times 10^{-3}$, respectively.

With trained TOPOS-VAE frozen, TOPOS-DiT learns a conditional flow matching model in its latent space. We use a DiT-style transformer backbone with model channels 1024, 24 blocks, 16 attention heads with head dimension 64 and an MLP expansion ratio of 4. The input and output channels match the TOPOS-VAE's latent dimension of 32, while cross-attention to the image condition uses 1024 channels. The image features are extracted by a frozen DINOv2 ViT-L/14-with-registers~\cite{oquab2023dinov2} backbone, producing 1374 patch tokens of dimension 1024, layer-normalized before being fed into the DiT. The timesteps are drawn from a logit-normal distribution with mean 0 and standard deviation 1, (i.e. $t = \sigma(\mathcal{N}(0,1))$). To enable classifier-free guidance, the image condition is dropped with probability 0.1 during training. The model is trained with AdamW~\cite{loshchilov2017decoupled} at a learning rate of $1\times 10^{-4}$ under the same polynomial schedule of TOPOS-VAE. At inference stage of TOPOS-DiT, we use a Flow Euler sampler with 50 steps and a CFG guidance scale of 3.0.

TOPOS-Texture fine-tunes Qwen-Image-Edit~\cite{wu2025qwenimagetechnicalreport} via LoRA with rank 32. The learning rate follows a cosine decay schedule from $5\times 10^{-5}$ to $3.5\times 10^{-5}$. We adopt a progressive resolution training strategy, starting from $256^2$ and gradually increasing the proportion of higher-resolution samples ($384^2$, $512^2$, $768^2$), until training exclusively on $1024^2$ images. The input condition is dropped with probability 0.2 for classifier-free guidance. All other hyperparameters follow the default settings of Qwen-Image-Edit~\cite{wu2025qwenimagetechnicalreport}. At inference stage of TOPOS-Texture, we apply a CFG guidance scale of 1.2.

\section{Geometric Augmentation Algorithm}
\label{sec:augmentation}

Our geometric augmentation algorithm is mainly implemented via region-wise blendshape sampling. We group the face into nine semantic regions: \{\textit{Brow, Eye, Nose, Mouth, Ear, Forehead, Cheekbone, FaceShape, Chin}\}.
For each base mesh $M$ and each facial region $r$, we randomly select three blendshapes
$\{B_{r,1}, B_{r,2}, B_{r,3}\}$ from the corresponding blendshape set, where $B_{r,i} \in \mathbb{R}^{V \times 3}$ denotes the $i$-th blendshape of region $r$, and $V$ is the total number of head mesh vertices. Their coefficients are independently sampled within artist-defined bounds,
\begin{equation}
w_{r,i} \sim \mathcal{U}(l_{r,i}, u_{r,i}), \quad i=1,2,3,
\end{equation}
where $l_{r,i}$ and $u_{r,i}$ denote the lower and upper bounds specified by artists. To avoid excessive deformation within a local region, we normalize the coefficients only when their sum exceeds $1$,
\begin{equation}
\tilde{w}_{r,i} =
\begin{cases}
\dfrac{w_{r,i}}{\sum_{j=1}^{3} w_{r,j}}, & \text{if } \sum_{j=1}^{3} w_{r,j} > 1,\\[6pt]
w_{r,i}, & \text{otherwise}.
\end{cases}
\end{equation}
The augmented head mesh $M'$ is then written as
\begin{equation}
M' = M + \sum_{r} \sum_{i=1}^{3} \tilde{w}_{r,i} B_{r,i}.
\end{equation}

All parameter ranges are specified by artists to suppress implausible facial shapes, and no additional manual inspection or post-filtering is applied, which enables efficient large-scale geometric augmentation.

\section{3D Head Mesh Alignment for Evaluation.}
\label{sec:alignment}
Since different methods output face meshes in heterogeneous coordinate frames, with varying scales, and with different spatial extents (some emit only the frontal face shell, while the ground-truth meshes cover the full head down to the neck and shoulders), a direct vertex-wise comparison is meaningless. Before computing any geometric metric, we therefore align each generated mesh $\mathcal{M}_P$ to its paired ground-truth mesh $\mathcal{M}_G$ by a 7-DoF similarity transform $(s, \mathbf{R}, \mathbf{t}) \in \mathbb{R}^{+}\times SO(3)\times \mathbb{R}^3$ so that $\mathbf{p}\mapsto s \mathbf{R}\mathbf{p}+\mathbf{t}$ maps the facial region of $\mathcal{M}_P$ onto that of $\mathcal{M}_G$. The ground-truth 3D head mesh is kept fixed throughout.

We solve for $(s,\mathbf{R},\mathbf{t})$ by trimmed similarity ICP~\cite{besl1992method, zinsser2005point,chetverikov2005robust}. At each iteration, every source vertex is matched to its nearest neighbor on $\mathcal{M}_G$ via a $k$-d tree, and only the top $\tau{=}50\%$ of correspondences with the smallest residuals are retained. A closed-form Umeyama estimate~\cite{umeyama2002least} is computed on this trimmed set and composed with the running transform. The trimming is essential: it automatically rejects correspondences that belong to regions present in one mesh but not the other (e.g., neck and shoulders in the ground truth, or a frontal-only shell in the prediction), so that the fit is dominated by the mutually-overlapping facial region rather than being dragged by extraneous geometry.

Because ICP is non-convex, we initialize it carefully. A coarse centroid translation and a bbox-diagonal-based scale $s_0$ provide a starting point. On top of this, we perform a small grid search over ${0^\circ,90^\circ,180^\circ,270^\circ}$ yaw rotations and over scale multipliers $\{0.5,\,0.7,\,1.0,\,1.4\}\cdot s_0$; the scale sweep is necessary because the bbox-diagonal heuristic is biased when the source covers only a subset of the target. For each candidate, a short 30-iteration trimmed ICP is run, and the one with the lowest residual seeds a final full-length refinement (up to 300 iterations, $\tau{=}50\%$).

The recovered transform is then applied to every vertex of $\mathcal{M}_P$, while connectivity, UVs, materials, and vertex normals are preserved verbatim (normals are rotated by $\mathbf{R}$ only and renormalized, since uniform scaling preserves their direction). For methods that ship their output 3D head meshes as GLB files, the same similarity is applied to the scene-graph root transform so that the full material/texture/node hierarchy remains intact. This yields a set of aligned meshes on which we compute all reported metrics.

To prevent extraneous geometry retained by some methods (e.g., neck and shoulders) from dominating the metrics, we clip $\mathcal{M}_G$ to the axis-aligned bounding box of $\mathcal{M}_R$, inflated by $5\%$, via exact plane-based triangle slicing. $\mathcal{M}_R$ itself is left intact, so that predictions failing to cover the face are still penalized through incompleteness. All Euclidean distances are normalized by the bounding-box diagonal of $\mathcal{M}_R$, making the F-score thresholds scale-free and comparable across different 3D head meshes.

\section{Evaluation Metrics Details}
\label{sec:metrics}

We denote the generated mesh and ground-truth mesh by $\mathcal{M}_G$ and $\mathcal{M}_R$, on which we randomly sample a set of N = 10k points $G = \{x_i\}_1^N$ and $R = \{y_i\}_1^N$, respectively. Define $\mathcal{P}_A(x) = \text{argmin}_{y\in A} ||x-y||$, which finds the closest point of $x$ from a point set $A$. The $L_1$ Chamfer distance is defined as:
\begin{equation}
    CD = \frac{1}{N}\sum_i\left\|x_i - \mathcal{P}_R(x_i)\right\| + \frac{1}{N} \sum_i \left\|y_i - \mathcal{P}_G(y_i)\right\|.
\end{equation} 

We define $\mathcal{N}(x)$ as an operator that returns the corresponding normal of an input point, then the normal consistency is define as:
\begin{equation}
    NC = \frac{1}{N}\sum_i\big|\mathcal{N}(x_i) \cdot \mathcal{N} \big(\mathcal{P}_R(x_i)\big)\big| + \frac{1}{N}\sum_i\big|\mathcal{N}(y_i) \cdot \mathcal{N}\big(\mathcal{P}_G(y_i)\big)\big|.
\end{equation}
The F-Score is defined as the harmonic mean between the precision and the recall of points that lie
within a certain distance between $\mathcal{M}_G$ and $\mathcal{M}_R$.

For 3D head texture evaluation, we calculate the perceptual similarity metric LPIPS~\cite{zhang2018unreasonable} based on AlexNet~\cite{krizhevsky2012imagenet} between the rendered images of generated head meshes and ground truth head meshes.

We also utilize FID~\cite{parmar2022aliased} and KID~\cite{binkowski2018demystifying} to compare the distribution of these two images sets. FID is defined as:
\begin{equation}
\text{FID}=\left\|\mu_G-\mu_R\right\|^2 + \text{Tr}\big(\Sigma_G + \Sigma_R - 2(\Sigma_R\Sigma_G)^{1/2}\big),
\label{equ:fid}
\end{equation}
and KID is defined as the squared Maximum Mean Discrepancy (MMD) between the two feature sets with a polynomial kernel $k(x,y)=\big(\tfrac{1}{d}x^\top y + 1\big)^{3}$, where $d$ is the feature dimension. Given features $\{g_i\}_{i=1}^{m}$ from $G$ and $\{r_j\}_{j=1}^{n}$ from $R$, KID is computed using the unbiased estimator:
\begin{equation}
\begin{aligned}
\text{KID} = \,& \frac{1}{m(m-1)}\sum_{i\neq i'}^{m} k(g_i,g_{i'}) 
+ \frac{1}{n(n-1)}\sum_{j\neq j'}^{n} k(r_j,r_{j'}) \\
& - \frac{2}{mn}\sum_{i=1}^{m}\sum_{j=1}^{n} k(g_i,r_j),
\end{aligned}
\label{equ:kid}
\end{equation}
where $G$ and $R$ denote the features of the generated images set and ground truth images set, which is extracted by Inception-v3 model~\cite{szegedy2016rethinking}. $\mu$ and $\Sigma$ denote the mean and covariance matrices of each image set.

Finally, we utilize Cosine SIMilarity of identity features (CSIM) to measure the identity preservation between two portrait images, which can be either rendered images or in-the-wild portrait images. We calculate CSIM through the cosine similarity of two embeddings from the representative pretrained face recognition network ArcFace~\cite{deng2019arcface}.

\section{More Generative Results}
\label{sec:results}
\begin{figure*}[t!]
    \includegraphics[width=\textwidth]{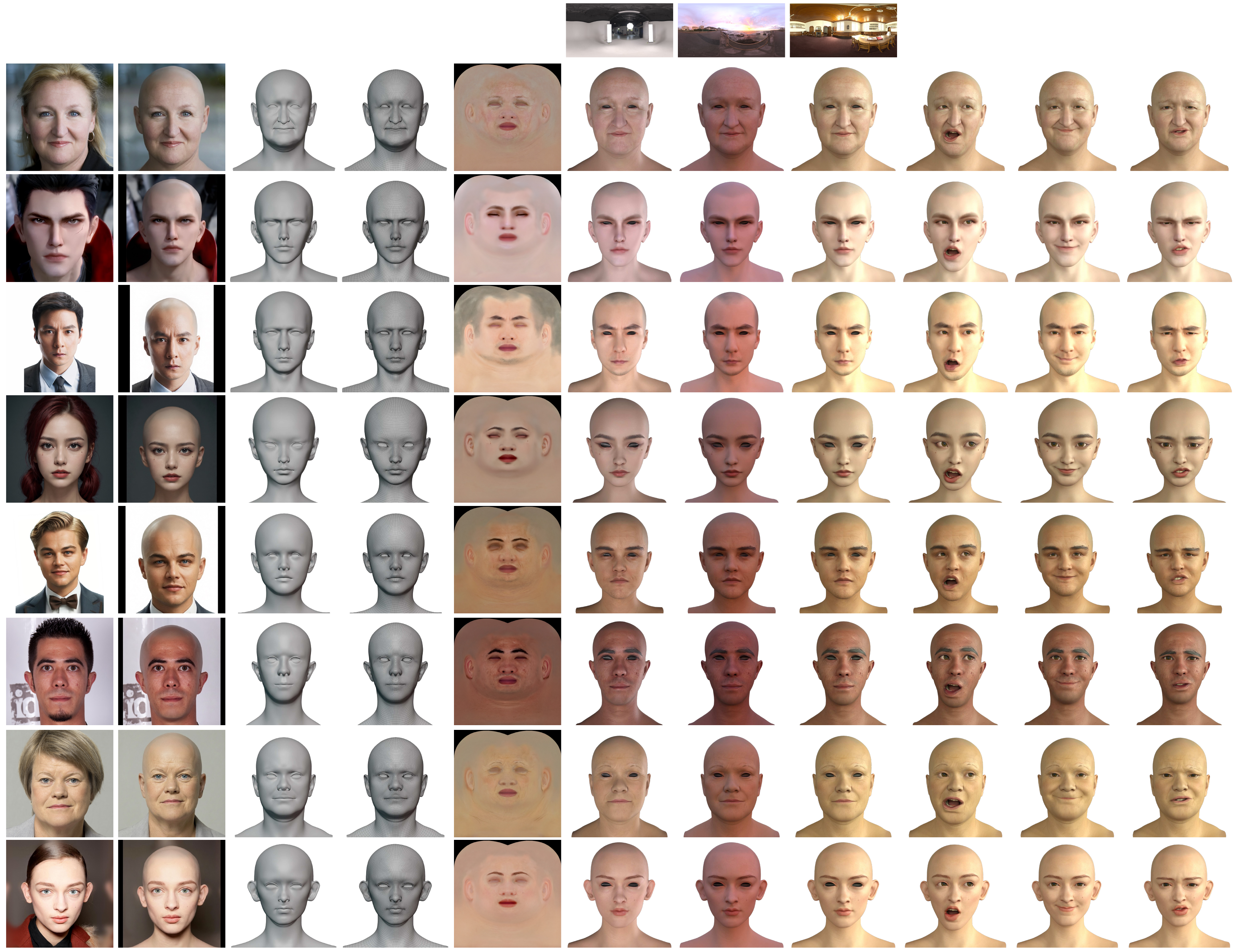}
    \vspace{-3mm}
    \caption{More generative results of our method on different in-the-wild images. From left to right are the input images, edited images, generated geometry, mesh topology, generated texture maps, rendering images under different lighting conditions and animation results across different expressions, respectively. The first row illustrates the environment maps. Please zoom in for better inspection.}
    \label{fig:results}
\end{figure*}

We present more 3D head results generated by our TOPOS framework in~\figref{fig:results}.

\section{Ethics Statement}
\label{sec:ethics}
This work advances single image conditioned 3D head generation. Our method is not intended for malicious use, and all synthesized content should clearly indicate its artificial nature. We acknowledge potential misuse, such as deepfakes, and are developing tools to help detect synthetic images and videos. At the same time, our technology can support education, communication assistance, and therapeutic applications, reflecting our commitment to responsible and ethical AI development.

\end{document}